\documentclass[sigconf]{acmart}

\AtBeginDocument{%
  \providecommand\BibTeX{{%
    \normalfont B\kern-0.5em{\scshape i\kern-0.25em b}\kern-0.8em\TeX}}}

\copyrightyear{2023}
\acmYear{2023}
\setcopyright{acmlicensed}
\acmConference[KDD'23]{Proceedings of the 29th ACM SIGKDD Conference on Knowledge Discovery and Data Mining}{August 6--10, 2023}{Long Beach, CA, USA}
\acmBooktitle{Proceedings of the 29th ACM SIGKDD Conference on Knowledge Discovery and Data Mining (KDD '23), August 6--10, 2023, Long Beach, CA, USA}
\acmPrice{15.00}
\acmDOI{10.1145/3580305.3599431}
\acmISBN{979-8-4007-0103-0/23/08}

\usepackage{mathtools}
\usepackage{amsthm}

\usepackage{multicol}
\usepackage{enumitem}

\theoremstyle{plain}
\newtheorem{theorem}{Theorem}[section]

\theoremstyle{definition}
\newtheorem{definition}[theorem]{Definition}

\theoremstyle{remark}

\usepackage{tikz}
\usetikzlibrary{positioning, arrows.meta}
\usetikzlibrary{decorations.pathmorphing, patterns}
\usepackage{mdframed}
\usepackage{threeparttable}
\usepackage[normalem]{ulem}
\usepackage{multirow}
\usepackage{makecell}
\usepackage{bbding}

\mdfsetup{skipabove=5pt,skipbelow=5pt}

\newcommand\bs[1]{\boldsymbol{#1}}

\newcommand\ul[1]{\underline{#1}}

\begin{document}

\title{MGNN: Graph Neural Networks Inspired by Distance Geometry Problem}


\author{Guanyu Cui}
\affiliation{%
  \institution{Renmin University of China}
  \city{Beijing}
  \country{China}}
\email{cuiguanyu@ruc.edu.cn}

\author{Zhewei Wei}
\authornote{Zhewei Wei is the corresponding author. The work was partially done at Gaoling School of Artificial Intelligence, Peng Cheng Laboratory, Beijing Key Laboratory of Big Data Management and Analysis Methods and MOE Key Lab of Data Engineering and Knowledge Engineering. }
\affiliation{%
  \institution{Renmin University of China}
  \city{Beijing}
  \country{China}}
\email{zhewei@ruc.edu.cn}


\begin{abstract}
Graph Neural Networks (GNNs) have emerged as a prominent research topic in the field of machine learning.
Existing GNN models are commonly categorized into two types: spectral GNNs, which are designed based on polynomial graph filters, and spatial GNNs, which utilize a message-passing scheme as the foundation of the model.
For the expressive power and universality of spectral GNNs, a natural approach is to improve the design of basis functions for better approximation ability.
As for spatial GNNs, models like Graph Isomorphism Networks (GIN) analyze their expressive power based on Graph Isomorphism Tests.
Recently, there have been attempts to establish connections between spatial GNNs and geometric concepts like curvature and cellular sheaves, as well as physical phenomena like oscillators.
However, despite the recent progress, there is still a lack of comprehensive analysis regarding the universality of spatial GNNs from the perspectives of geometry and physics.

In this paper, we propose MetricGNN (MGNN), a spatial GNN model inspired by the congruent-insensitivity property~\footnote{That is, they can produce identical outputs when given two congruent input matrices. Congruent matrices are defined as matrices where the distance between each pair of corresponding rows remains the same. Formally, for every pair of $i$ and $j$, the equality $d(\bs{A}_{i:}, \bs{A}_{j:}) = d(\bs{B}_{i:}, \bs{B}_{j:})$ holds.} of classifiers in the classification phase of GNNs. 
We demonstrate that a GNN model is universal in the spatial domain if it can generate embedding matrices that are congruent to any given embedding matrix. This property is closely related to the Distance Geometry Problem (DGP).
Since DGP is an NP-Hard combinatorial optimization problem, we propose optimizing an energy function derived from spring networks and the Multi-Dimensional Scaling (MDS) problem.
This approach also allows our model to handle both homophilic and heterophilic graphs. 
Finally, we propose employing the iteration method to optimize our energy function. We extensively evaluate the effectiveness of our model through experiments conducted on both synthetic and real-world datasets.
Our code is available at: \url{https://github.com/GuanyuCui/MGNN}.
\end{abstract}

\ccsdesc[500]{Computing methodologies~Neural networks}
\ccsdesc[500]{Computing methodologies~Machine learning algorithms}

\keywords{graph neural networks, distance geometry problem, metric matrix}



\maketitle

\section{Introduction}
\paragraph{Graph Neural Networks.} 
With the rapid growth of graph data, Graph Neural Networks (GNNs) have been widely used in many areas such as physics simulation~\cite{sanchezgonzalez2020learning}, traffic forecasting~\cite{li2018diffusion}, and recommendation systems~\cite{ying2018graph, wu2019session}, etc..
A typical scheme of a GNN consists of following components:
\begin{itemize}
	\item An embedding function $f_\theta(\cdot)$ which maps the original node feature matrix $\bs{X}\in \mathbb{R}^{n\times F}$ to an initial embedding matrix $\bs{Z}^{(0)}$. 
		The linear layers and MLPs are commonly used both in spatial models, such as GCNII~\cite{chen2020simple}, and spectral models, such as ChebNet~\cite{defferrard2016convolutional}, BernNet~\cite{he2021bernnet}, etc..
	\item A graph propagation / convolution module $GP(\bs{Z}; \bs{A})$ which propagates the hidden features $\bs{Z}$ on the graph with adjacency matrix $\bs{A}$.
		While spectral models are inspired by spectral graph theory, it is worth noting that the filter utilized in these models can also be implemented using the message passing method.
	\item An optional pooling or readout module $R(\bs{Z}^{(K)}; \bs{A})$ which extract features from the final embedding matrix $\bs{Z}^{(K)}$ for graph-level tasks.
	\item A final output function $g_\theta(\bs{Z}^{(K)})$ for classification or regression with the embeddings generated. The linear layers and MLPs are also commonly used.
\end{itemize}
We can refer to the initial two stages as the embedding phase of the GNN, while the subsequent stages can be referred to as the classification phase~\footnote{Some regression tasks also use MLPs as their regression functions, we can also regard them as classification tasks.}. 
Figure \ref{fig:gnn-scheme} is an illustration of a typical scheme of GNNs.

\begin{figure}[ht]
	\centering
	\resizebox{\linewidth}{!}{
	\begin{tikzpicture}
		\node[fill = gray!30] at (-1, 0){$\bs{X}$};
		\draw[->, thick] (-.7, 0) -- (-.2, 0);
		\draw[thick, fill = teal!50, draw opacity = 1] (0, 1) -- (0, -1) -- (1, -.5) -- (1, .5) -- cycle;
		\node at (.5, 0) {$f_\theta(\cdot)$};
		\node[fill = cyan!30] at (.5, 1.5) {Emb. Function};
		\draw[->, thick] (1.2, 0) -- (1.8, 0);

		\draw[thick, fill = orange!50, rounded corners = 2mm, draw opacity = 1] (2, 1.5) rectangle (7, -1.5);
		\node[above, fill = cyan!30] at (4.5, 1.7){Graph Prop. / Conv. Layers};
		\begin{scope}[xshift = 1.85cm]
			\draw[fill = gray!20, fill opacity = .8, rounded corners = 2mm, draw opacity = .0] (.4, 1.) rectangle (1.7, -1.);
			\draw (1., 0) -- (.6, -.5);
			\draw (1., 0) -- (.7, .4);
			\draw (1., 0) -- (1.5, 0);
			\draw (.6, -.5) -- (.7, .4);
			\draw (1.5, .0) -- (1.3, -.4);
			\draw[fill = blue!50, draw opacity = 1] (1., 0) circle (.1cm);
			\draw[fill = red!50, draw opacity = 1] (.6, -.5) circle (.1cm);
			\draw[fill = teal!50, draw opacity = 1] (.7, .4) circle (.1cm);
			\draw[fill = yellow!50, draw opacity = 1] (1.5, 0) circle (.1cm);
			\draw[fill = green!50, draw opacity = 1] (1.3, -.4) circle (.1cm);
		\end{scope}
		\draw [->, thick] (3.65, 0) -- (4.15, 0);
		\begin{scope}[xshift = 3.85cm]
			\draw[thick, fill = gray!20, fill opacity = .8, rounded corners = 2mm, draw opacity = .0] (.4, 1.) rectangle (1.7, -1.);
			\draw (1., 0) -- (.6, -.5);
			\draw (1., 0) -- (.7, .4);
			\draw (1., 0) -- (1.5, 0);
			\draw (.6, -.5) -- (.7, .4);
			\draw (1.5, .0) -- (1.3, -.4);
			\draw[fill = blue!50, draw opacity = 1] (1., 0) circle (.1cm);
			\draw[fill = red!50, draw opacity = 1] (.6, -.5) circle (.1cm);
			\draw[fill = teal!50, draw opacity = 1] (.7, .4) circle (.1cm);
			\draw[fill = yellow!50, draw opacity = 1] (1.5, 0) circle (.1cm);
			\draw[fill = green!50, draw opacity = 1] (1.3, -.4) circle (.1cm);
		\end{scope}
		\draw [->, thick] (5.65, 0) -- (6.15, 0) node[right = .1cm] {$\cdots$};

		\draw [->, thick] (7.2, 0) -- (7.7, 0);
		\node[fill = gray!30] at (8, 0){$\bs{Z}$};

		\draw[->, dashed] (8., .4) -- (8., 1.5) -- (8.8, 1.5);
		\draw[thick, fill = blue!50, draw opacity = 1.] (9, 2.5) -- (9, 0.5) -- (10, 1) -- (10, 2) -- cycle;
		\node at (9.5, 1.5) {$R(\cdot)$};
		\node[fill = cyan!30] at (9.5, 3) {Pooling / Read-Out};

		\draw [->, thick] (8.4, 0) -- (10.8, 0);
		\draw[thick, fill = teal!50, draw opacity = 1] (11, 1) -- (11, -1) -- (12, -.5) -- (12, .5) -- cycle;
		\draw[->, dashed] (10.2, 1.5) -- (11.5, 1.5) -- (11.5, 1);
		\node at (11.5, 0) {$g_\theta(\cdot)$};

		\node[fill = cyan!30] at (11.5, -1.5) {Class. Function};

		\draw[->, thick] (12.2, 0) -- (12.7, 0);
		\node[fill = gray!30] at (13, 0){$\bs{Y}$};

		\draw [dashed] (7.5, -2.5) -- (7.5, 3.5);
		\node[left = .1cm] at (7.5, -2.2){Embedding Phase $\gets$};
		\node[right = .1cm] at (7.5, -2.17){$\to$ Classification Phase};
	\end{tikzpicture}
	}
	\caption{A typical scheme of GNNs.}
	\label{fig:gnn-scheme}
\end{figure}

\paragraph{Expressive Power and Universality of GNNs.}
There is a significant body of literature dedicated to exploring the expressive power and universality of GNNs.
Notably, the GIN model~\cite{xu2019how} is the first to highlight the connection between the expressive power of GNNs and the Weisfeiler-Lehman graph isomorphism test.
Furthermore, Chen et al.~\cite{chen2019on} establishes the equivalent relation between graph isomorphism test and function approximation on graphs.
There are works trying to discuss the design of universal GNNs from the spectral perspective as well.
For instance, Balcilar et al.~\cite{balcilar2021analyzing} bridges the gap between spectral and spatial perspectives by reformulating various GNNs within a common framework.
BernNet~\cite{he2021bernnet} proposes to use Bernstein basis to approximate arbitrary graph spectral filter, and Wang et al.~\cite{wang2022how} proves that linear GNNs can generate arbitrary graph signals, under some conditions regarding the graph Laplacian $\bs{L}$ and node features $\bs{X}$.

\paragraph{Our Contribution.}
In this paper, we discuss the universality of GNNs from a different geometric and physical perspective.
Our contributions can be summarized as follows:
\begin{itemize}
	\item Inspired by the congruent-insensitivity property of the classification phase of GNNs, we propose that a spatial-universal GNN should possess the capability to generate congruent embeddings with any embedding matrix produced by other GNNs.
	\item To generate congruent embeddings, we propose the MGNN model, which aims to minimize an energy function that carries physical meaning. We also discuss the relation of that objective function to the distance geometry problem, the multidimensional scaling problem, which provides strong theoretical backgrounds for our model.
	\item We propose to use the stationary point iteration method to optimize the objective function.
	\item We design experiments with synthetic and real-world graphs to show the effectiveness of the proposed MGNN model.
\end{itemize}

\section{Other Related Works}
In this section, we will list some other related works.
\paragraph{Distance Geometry and Graph Rigidity.}
Distance Geometry (DG) is a mathematical discipline that focuses on determining the spatial arrangement of points by utilizing given pairwise distance constraints as input. 
It has applications in various fields, including sensor network localization~\cite{biswas2006semidefinite}, and molecular conformation analysis~\cite{crippen1988distance}, etc..
The core problem in DG is known as the following Distance Geometry Problem (DGP)~\cite{liberti2014euclidean, liberti2017euclidean}:
\begin{mdframed}[backgroundcolor = gray!30, frametitle = Distance Geometry Problem (DGP)]
	Given an integer $d>0$, a simple undirected graph $G = (V, E)$, and a symmetric non-negative metric matrix $\bs{M}$, decide whether there exists an embedding matrix $\bs{Z}\in \mathbb{R}^{\vert V\vert \times d}$, such that 
	$$\forall (i, j)\in E,  \Vert \bs{Z}_{i:} - \bs{Z}_{j:}\Vert = \bs{M}_{ij}.$$
	When the norm $\Vert\cdot\Vert$ is the Euclidean norm, this problem is also called Euclidean Distance Geometry Problem (EDGP).
\end{mdframed}
Graph Rigidity refers to the property of a graph to maintain its shape or configuration when subjected to external forces. 
If we imagine connecting the nodes of a graph with bars, the graph is considered rigid if its shape remains unchanged regardless of the movements or repositioning of its nodes~\cite{alfakih2007dimensional}.

\paragraph{Geometric- and Physics-Inspired GNNs.}
There has been a growing interest in exploring the connections between GNNs and various geometric objects. 
For instance,~\cite{topping2022understanding} generalizes curvature for graphs and reveals the relation between over-squashing phenomenon and graph curvature. 
Furthermore, there are other works utilize gradient flow~\cite{giovanni2022graph} and cellular sheaves on graphs~\cite{bodnar2022neural} to design new GNN structures. 
Additionally, there has been research focusing on the connections between GNNs and physical concepts.
For example,~\cite{rusch2022graph} explores the relation between GNNs and oscillators, while~\cite{wang2023acmp} investigates the potential energy concepts from particle physics in the context of GNNs.

\paragraph{Optimization Derived GNNs.}
Several works focus on deriving new GNNs from carefully designed optimization objectives (usually energy functions) or consolidating existing models into a unified optimization framework.
For instance, works such as~\cite{yang2021graph, ma2021unified} propose new GNN models based on the graph signal denoising objective function.
This objective function can be formulated as the sum of an $\ell_2$ regularization term and a Laplacian regression term:
\begin{equation}
    \mathcal{L}(\bs{z}) = \Vert \bs{z} - \bs{x}_{\text{in}}\Vert^2_2 + \lambda \bs{z}^\top\bs{L}\bs{z}.
\end{equation}
Subsequent works have extended and generalized the graph signal denoising optimization objective in various ways. 
For example, BernNet~\cite{he2021bernnet} generalizes the Laplacian matrix in the regression term to any positive semidefinite function of the Laplacian.
ElasticGNN~\cite{liu2021elastic} introduces additional $\ell_1$ and $\ell_{21}$ regularization terms into the objective function.
$^p$GNN~\cite{fu2022p} generalizes the Laplacian matrix to the $p$-Laplacian matrix, and ACMP-GNN~\cite{wang2023acmp} both modify the Laplacian matrix and replace the $\ell_2$ regularization term with the double-well potential. 

\section{Preliminaries}
\paragraph{Notations for Matrices and Vectors.}
We use boldface uppercase letters such as $\bs{X}$ for matrices; while boldface lowercase characters such as $\bs{x}$ denote column vectors. 
For matrices, $\bs{X}_{i:}$ denotes the $i$-th row of $\bs{X}$, $\bs{X}_{:j}$ denotes the $j$-th column of $\bs{X}$, $\bs{X}_{ij}$ denotes the $ij$-th element of $\bs{X}$, and $\bs{X}^\top$ denotes the transposition of $\bs{X}$. 
Meanwhile, $\mathrm{diag}(\bs{X})$ denotes the column vector that consists of all diagonal elements of $\bs{X}$, i.e., $\mathrm{diag}(\bs{X})_i=\bs{X}_{ii}$ ($i=1, 2, \cdots, n$).
For vectors, $\bs{x}_i$ denotes the $i$-th element of $\bs{x}$, whereas $\bs{x}^\top$ denotes the transposition of $\bs{x}$.
And $\mathrm{diag}(\bs{x})$ denotes the square matrix whose diagonal elements are $\bs{x}$, i.e., $\mathrm{diag}(\bs{x})_{ij}=\begin{cases}\bs{x}_i, & i=j \\ 0, & i\neq j\end{cases}$.
The Hadamard product is the element-wise product of two matrices or vectors, i.e., $(\bs{X}\odot \bs{Y})_{ij}=\bs{X}_{ij}\bs{Y}_{ij}$, and $(\bs{x}\odot\bs{y})_i=\bs{x}_i\bs{y}_i$.

\paragraph{Notations From the Graph Theory.}
An (unweighted) graph $G$ is a pair $(V, E)$, where $V$ is the node set and $E\subseteq V^2$ is the edge set.
We always denote $\vert V\vert$ and $\vert E\vert$ as $n=\vert V\vert$ and $m = \vert E\vert$.
Given an arbitrary order for the nodes, we may denote $V$ as $\{1, 2, \cdots, n\}$, and use the adjacency matrix $\bs{A}\in \{0, 1\}^{n\times n}$ to represent the edges $E$, i.e., $\bs{A}_{ij}=1$ if and only if edge $(i, j)\in E$, otherwise $\bs{A}_{ij}=0$. 
Note that for an undirected graph, $(i, j)\in E\Leftrightarrow (j, i)\in E$, resulting in a symmetric $\bs{A}$. 
Unless otherwise noted, the graphs below are undirected.
The neighborhood of node $i$ is the set of nodes that share an edge with it, i.e., $\mathcal{N}(i)\coloneqq\{j : (i, j)\in E\}$. 
The degree of node $i$ is the number of nodes in its neighborhood, i.e., $d_i\coloneqq\vert\mathcal{N}(i)\vert=\vert \{j : (i, j)\in E\}\vert$.
And the degree matrix of graph is $\bs{D}\coloneqq \mathrm{diag}((d_1, d_2, \cdots, d_{n}))$.
The Laplacian of graph is defined as $\bs{L}\coloneqq\bs{D}-\bs{A}$. 

\paragraph{Concepts From the Graph Rigidity Theory.}
We list the definitions of \textit{equivalent}, \textit{congruent}, \textit{rigid} and \textit{globally rigid} in the graph rigidity theory~\cite{jackson2007notes}:
\begin{definition}[Equivalent]
    Two node embedding matrices $\bs{Z}^{(1)}$ and $\bs{Z}^{(2)}\in \mathbb{R}^{n\times d}$ of a graph $G$ are equivalent (denoted as $\bs{Z}_1\equiv_E \bs{Z}_2$) if $\Vert \bs{Z}^{(1)}_{i:}-\bs{Z}^{(1)}_{j:}\Vert_2=\Vert \bs{Z}^{(2)}_{i:}-\bs{Z}^{(2)}_{j:}\Vert_2$ for all $(i, j)\in E$.
\end{definition}
       
\begin{definition}[Congruent]
    Two node embedding matrices $\bs{Z}^{(1)}$ and $\bs{Z}^{(2)}\in \mathbb{R}^{n\times d}$ of a graph $G$ are congruent (denoted as $\bs{Z}_1\cong_{V^2} \bs{Z}_2$) if $\Vert \bs{Z}^{(1)}_{i:}-\bs{Z}^{(1)}_{j:}\Vert_2=\Vert \bs{Z}^{(2)}_{i:}-\bs{Z}^{(2)}_{j:}\Vert_2$ for all $i, j\in V$.
\end{definition}

\begin{definition}[Globally Rigid]
    An embedding matrix $\bs{Z}$ of a graph $G$ is globally rigid if all its equivalent embedding matrices $\bs{Z}'$ are also congruent to $\bs{Z}$.
\end{definition}

\begin{definition}[Rigid]
    An embedding matrix $\bs{Z}$ of a graph $G$ is rigid if there exists $\varepsilon>0$ such that every equivalent embedding $\bs{Z}'$ which satisfies $\Vert \bs{Z}_{v:}-\bs{Z}'_{v:}\Vert_2< \varepsilon$ for all $v\in V$, is congruent to $\bs{Z}$. 
		Or informally, all equivalent embeddings that can be obtained by \textbf{continuous} motion from $\bs{Z}$ are congruent to $\bs{Z}$.
\end{definition}

It is important to note that the globally rigid condition is stronger than the rigid condition because not all rigid graphs are globally rigid. 
For instance, consider the following graph in $\mathbb{R}^2$. 
It is rigid because any continuous motion applied to it will lead to a set of equivalent embeddings.
However, there exists an equivalent embedding that is not congruent.
\begin{figure}[ht]
    \centering
    \begin{tikzpicture}
        \draw (0, 0) -- (1, 0) -- (0.5, 0.866) -- cycle;
        \draw (0, 0) -- (0.5, 0.289) -- (1, 0);
        \draw[fill = white] (0, 0) circle (.1);
        \draw[fill = white] (1, 0) circle (.1);
        \draw[fill = white] (0.5, 0.866) circle (.1);
        \draw[fill = white] (0.5, 0.289) circle (.1);

        \begin{scope}[xshift = 3cm]
            \draw (0, 0) -- (1, 0) -- (0.5, 0.866) -- cycle;
            \draw (0, 0) -- (0.5, -0.289) -- (1, 0);
            \draw[fill = white] (0, 0) circle (.1);
            \draw[fill = white] (1, 0) circle (.1);
            \draw[fill = white] (0.5, 0.866) circle (.1);
            \draw[fill = white] (0.5, -0.289) circle (.1);
        \end{scope}
    \end{tikzpicture}
    \caption{A rigid but not globally rigid graph (left) in $\mathbb{R}^2$, since it has an equivalent but not congruent embedding (right).}
\end{figure}

We also define the equivalent and congruent transformations and their corresponding groups for the convenience of discussion:
\begin{definition}[Equivalent Transformation]
	Given a graph $G=(V, E)$ and an embedding matrix $\bs{Z}$, a reversible transformation $\mathcal{T}:\mathbb{R}^{n\times d}\to \mathbb{R}^{n\times d}$ is called $E$-equivalent if $\mathcal{T}(\bs{Z})\equiv_E \bs{Z}$ always holds for all $\bs{Z}\in \mathbb{R}^{n\times d}$ .
\end{definition}
\begin{definition}[Congruent Transformation]
	Given a graph $G = (V, E)$ and an embedding matrix $\bs{Z}$, a reversible transformation $\mathcal{T}:\mathbb{R}^{n\times d}\to \mathbb{R}^{n\times d}$ is called $V^2$-congruent if $\mathcal{T}(\bs{Z})\cong_{V^2} \bs{Z}$ always holds for all $\bs{Z}\in \mathbb{R}^{n\times d}$.
\end{definition}
It can be demonstrated straightforwardly that these two types of transformations constitute two groups, with transformation composition serving as the group operation. 
We denote them as $\mathrm{Iso}(n, d; E)$ and $\mathrm{Iso}(n, d; V^2)$, or simply as $\mathrm{Iso}(E)$ and $\mathrm{Iso}(V^2)$ when the matrix dimensions can be inferred from the context.
We have the following Theorem that character the relation between these groups:
\begin{theorem}
	\label{thm:groups-relation}
	$\mathrm{E}(d)\cong \mathrm{Iso}(n, d; V^2)\le \mathrm{Iso}(n, d; E)$, where $\mathrm{E}(d)$ is the $d$-dimensional Euclidean group, 
		i.e., the group containing the composition of orthogonal and translation transformations; $\cong$ denotes the group isomorphism relation, and $\le$ denotes the subgroup relation.
\end{theorem}
The proof of this theorem can be found in Appendix \ref{app:prf-group-relation}.

For the convenience of discussion, we also provide the definition of the metric matrix:
\begin{definition}[The Metric Matrix of an Embedding Matrix]
	The metric matrix of an embedding matrix $\bs{Z} \in \mathbb{R}^{n \times d}$ is defined as a matrix containing all pairwise distances between the embedding vectors, i.e., $(\bs{M}_{\bs{Z}})_{ij}=\Vert\bs{Z}_{i:}-\bs{Z}_{j:}\Vert_2$.
	The equivalent form is 
	\begin{equation}
    \label{eqn:metric-matrix}
	    \bs{M}_{\bs{Z}}=\left(\mathrm{diag}(\bs{ZZ}^\top)\bs{1}_n^\top+\bs{1}_n\mathrm{diag}(\bs{ZZ}^\top)^\top - 2\bs{ZZ}^\top\right)^{\odot \frac12},
	\end{equation} 
	where $\bs{X}^{\odot k}$ is the element-wise power of $\bs{X}$.
	We also define the mapping from an embedding matrix $\bs{Z}$ to its metric matrix $\bs{M}_{\bs{Z}}$ as $\bs{M}_{\bs{Z}} = M(\bs{Z})$.
\end{definition}

\section{Motivation}
\label{sec:motivation}
Several studies have examined the expressive power of GNNs.
Some of these works investigate the connection between message-passing GNNs and the Weisfeiler-Lehman (WL) graph isomorphism test, including~\cite{xu2019how, morris2019weisfeiler, maron2019provably, balcilar2021analyzing}.
Other studies focus on the relationship between spectral GNNs and graph filters, as explored in research papers such as~\cite{kipf2017semi, bo2021beyond, chien2021adaptive, he2021bernnet, wang2022how}.

Recently, there has been a growing trend of exploring the relationships between GNNs and geometric objects such as curvature~\cite{topping2022understanding}, gradient flow~\cite{giovanni2022graph}, and cellular sheaves on graphs~\cite{bodnar2022neural}.
Additionally, there is increasing interest in investigating the relationships between GNNs and physical objects such as oscillators~\cite{rusch2022graph} and potential energy in particle physics~\cite{wang2023acmp}.
To the best of our knowledge, there has been no previous research that analyzes the universality of spatial GNNs in relation to the metric matrix of embeddings and the distance geometry problem.
Our work will explore how to design spatial-universal GNNs from those aspects and will also reveal their tight connections to the graph rigidity, the multidimensional scaling problem, and the spring networks.

\subsection{Sufficient Conditions of a Spatial-Universal GNN}
Intuitively, we can observe that common classification modules in GNNs can yield identical predictions when given congruent embedding matrices.
In other words, applying congruent transformations to the embedding matrix does not impact the accuracy of predictions.
Formally, we can state the following theorem to capture this observation:
\begin{theorem}[MLPs Are Congruent-Insensitive]\label{thm:mlp-congruent-insensitive}
	Given two congruent embedding matrices $\bs{Z}_1$ and $\bs{Z}_2$, for any $\mathrm{MLP}_M$ (with bias)
	\footnote{MLP layers are formulated as $\bs{X}^{(k+1)}=\sigma\left(\bs{X}^{(k)}\bs{W}^{(k)}+\bs{1}(\bs{b}^{(k)})^\top\right)$. 
	Linear layers can be regarded as the special cases of MLP layers when $\sigma$ is the identity mapping.} 
	, there always exists another $\mathrm{MLP}_{N}$ (also with bias) such that $\mathrm{MLP}_{M}(\bs{Z}_1) = \mathrm{MLP}_{N}(\bs{Z}_2)$.
	That is, they produce identical predictions with $\bs{Z}_1$ and $\bs{Z}_2$, respectively.
\end{theorem}
The proof of this theorem can be found in Appendix \ref{app:prf-mlp-congruent-insensitive}.

Now we try to define the concept of a spatial-universal GNN. 
Universality refers to a model's capability to approximate various other models. 
A widely recognized universal model is the Universal Turing Machine (UTM)~\cite{turing1936computable}, which was introduced by the mathematician and computer scientist Alan Turing.
Similar to the idea of the Universal Turing Machine, which has the ability to simulate any Turing Machine (TM), we anticipate that a universal GNN $U$ can ``simulate'' any given GNN $A$ by generating identical predictions to those of $A$.
Based on Theorem \ref{thm:mlp-congruent-insensitive}, we can infer that if a GNN $U$ can generate an embedding matrix that is congruent to the one produced by any given GNN $A$ during its embedding phase, then $U$ is capable of generating identical predictions to those of $A$.

\subsection{Metric Matrix as a Guide}
Just like every programmable TM can function as a UTM, we can consider the metric matrix as the ``program'' or the guide to design a spatial-universal GNN.
If a GNN $U$ can generate an embedding matrix $\bs{Z}_U$ using the provided metric matrix $\bs{M}=M(\bs{Z}_{N})$ of any other GNN $N$, then $U$ is capable of producing identical predictions to $N$\footnote{To be precise, suppose $\bs{Z}^* = \mathrm{GNN}_N(\bs{X}; \bs{A})$, we can set $\bs{M}_{ij} = \Vert \bs{Z}^*_{i:} - \bs{Z}^*_{j:}\Vert_2$. In some cases, particularly for complex GNN models that we need to simulate, it may not be feasible to calculate the closed-form metric in advance. However, the theoretical existence of such metrics is still valid based on the aforementioned construction.}.
Given the observation that two embedding matrices are congruent if and only if their metric matrices are identical, we can infer that $\bs{Z}_U$ is congruent to $\bs{Z}_N$, indicating that $U$ is capable of generating identical predictions to $N$. 
Consequently, we can refer to $U$ as a spatial-universal GNN to a certain degree. 
An illustration of the concept of spatial-universal GNNs is presented in Figure \ref{fig:idea}.
\begin{figure}[ht]
	\centering
	\resizebox{\linewidth}{!}{
	\begin{tikzpicture}
		\node[fill = gray!30] at (0, 0) {$\bs{X}$};
		\draw[fill = orange!50, draw opacity = .5, rounded corners] (1, 0.5) rectangle node {Emb. Phase of $\mathrm{GNN}_N$} (5, 1.5);
		\node[fill = gray!30] at (6, 1) {$\bs{Z}_N$};
		\draw[fill = orange!50, draw opacity = .5, rounded corners] (7, 0.5) rectangle node {Class. Phase of $\mathrm{GNN}_N$} (11, 1.5);

		\draw[fill = orange!50, draw opacity = .5, rounded corners] (1, -0.5) rectangle node {Emb. Phase of $\mathrm{GNN}_U$} (5, -1.5);
		\node[fill = gray!30] at (6, -1) {$\bs{Z}_U$};
		\draw[fill = orange!50, draw opacity = .5, rounded corners] (7, -0.5) rectangle node {Class. Phase of $\mathrm{GNN}_U$} (11, -1.5);

		\node[fill = gray!30] at (12, 1) {$\bs{Y}_N$};
		\node[fill = gray!30] at (12, -1) {$\bs{Y}_U$};

		\draw [->] (0, .3) -- (0, 1) -- (.8, 1);
		\draw [->] (0, -.3) -- (0, -1) -- (.8, -1);
		\draw [->] (5.15, 1) -- (5.55, 1);
		\draw [->] (5.15, -1) -- (5.55, -1);
		\draw [->] (6.45, 1) -- (6.85, 1);
		\draw [->] (6.45, -1) -- (6.85, -1);
		\draw [->] (11.15, 1) -- (11.5, 1);
		\draw [->] (11.15, -1) -- (11.5, -1);

		\draw [->] (3, -2) node[right] {Some ``programs'' such as the given $M(\bs{Z}_N)$ or a learned $\bs{M}$.} -- (2, -2) -- (2, -1.5);

		\draw [<->, dashed] (6, .6) -- node[right]{$\bs{Z}_N\cong_{V^2}\bs{Z}_U$} (6, -.6);
		\draw [<->, dashed] (9, .5) -- node[right]{Theorem \ref{thm:mlp-congruent-insensitive}} (9, -.5);
		\draw [<->, dashed] (12, .6) -- node[right]{$\bs{Y}_N=\bs{Y}_U$} (12, -.6);
	\end{tikzpicture}
	}
	\caption{The idea of the spatial-universal GNNs.}
	\label{fig:idea}
\end{figure}

\section{The MGNN Model}
\label{sec:model}
\subsection{The Optimization Objective}
We have established that a GNN $U$ can be regarded as a spatial-universal GNN if it is capable of producing an embedding matrix which is congruent to the embedding matrix generated by any other GNN. 
Consequently, we anticipate that spatial-universal GNNs have the potential to tackle the Distance Geometry Problem (DGP)~\cite{liberti2017euclidean} with $E=V^2$.
Since the full metric matrix consists of all pairwise distances between the embedding vectors of nodes, computing the complete $\bs{M}$ matrix is a computationally intensive $O(n^2)$ operation.
To alleviate the heavy computational burden, we propose relaxing the constraints by solely considering node pairs connected by edges.
By doing so, computing the partial metric matrix becomes an $O(m)$ operation, which will lead to a local message-passing operation on the graph (see Section \ref{subsec:solution}).

This adaptation has resulted in a reduction of expressive power from generating congruent embeddings to generating equivalent embeddings with the given $\bs{M}$. 
In the case of a globally rigid graph, where the pairwise distances between nodes connected by edges suffice to determine the ``shape'' of the embedding, all equivalent embeddings of such a graph are congruent. 
Therefore, this adaptation does not weaken the expressive power of the model in such cases.
For a graph that is not globally rigid, its equivalent embeddings may not necessarily be congruent, and this adaptation weakens the expressive power in such cases. 
Unfortunately, testing the global rigidity of a graph is an NP-Hard problem~\cite{saxes1979embeddablity}, making it infeasible to test the global rigidity of the graph beforehand.
Besides, not all non-negative symmetric matrices are valid metric matrices.
For example, consider a triangle $K_3$ in $\mathbb{R}^2$, whose nodes are labelled as $V = \{1, 2, 3\}$.
$\bs{M}^{(1)} = 
\small{
\begin{bmatrix}
	0 & 3 & 4 \\
	3 & 0 & 5 \\
	5 & 4 & 0
\end{bmatrix}
}
$ is a valid matrix, as there exists an embedding matrix 
$\bs{Z} =
\begin{bmatrix}
	0 & 0 & 4 \\
	0 & 3 & 0 \\
\end{bmatrix}^\top
$ satisfying the metric constraints.
However, 
$\bs{M}^{(2)} = 
\small{
\begin{bmatrix}
	0 & 1 & 1 \\
	1 & 0 & 3 \\
	1 & 3 & 0
\end{bmatrix}
}
$ is not a valid matrix, since it does not satisfy the triangle inequality $\bs{M}_{1, 2} + \bs{M}_{1, 3}\ge \bs{M}_{2, 3}$.
Given that the DGP is NP-Hard~\cite{saxes1979embeddablity}, we can not effectively determine whether a given non-negative symmetric matrix is valid or not.

Hence, it becomes necessary to introduce an error-tolerant optimization objective for our model. 
We utilize the sum of squared estimate of errors (SSE) as the objective function:
\begin{align}
\begin{split}
	E_p(\bs{Z}; \bs{M}, E) &= \frac{1}{2}\Vert\bs{A}\odot\left(M(\bs{Z})-\bs{M}\right)\Vert_F^2 \\
	& =\sum_{(i, j)\in E} \frac{1}{2}\left(\Vert \bs{Z}_{i:} - \bs{Z}_{j:}\Vert_2 - \bs{M}_{ij}\right)^2.
\end{split}
\end{align}
This optimization objective is originated from the raw Stress function $\sigma_r$ in the Multidimensional Scaling (MDS) problem~\cite{kruskal1964multi, borg2005modern}.
Remarkably, this objective function also carries a physical interpretation.
Spring network models, which are a special case of force-directed methods, have been employed in the field of graph drawing since the 1960s~\cite{tutte1963how, kamada1989an, fruchterman1991graph, tunkelang1994a, gansner2004graph, wang2017revisiting}. 
These models have found applications in the biophysics and biochemistry domains for analyzing the properties of proteins~\cite{haliloglu1997gaussian, lin2009generalized, amyot2019analyzing}.
Suppose that we have a spring network with the underlying graph $G = (V, E)$ in Figure \ref{fig:spring-network}. 
By considering the edges $E$ as springs with characteristic constants $k_{ij} = 1$, the embedding matrix $\bs{Z}$ as the instantaneous positions of the nodes, and the metric matrix $\bs{M}$ as the relaxed lengths of the springs, the objective function $E_p(\bs{Z}; \bs{M}, E)$ represents the total elastic potential energy of the spring network.
\begin{figure}[ht]
\centering
    \resizebox{0.2\linewidth}{!}{
    \begin{tikzpicture}
        \draw [decoration={aspect=0.5, segment length=1.4mm, amplitude=.5mm, coil}, decorate] (0, 0) -- (1, 1.5);
        \draw [decoration={aspect=0.5, segment length=1.4mm, amplitude=.5mm, coil}, decorate] (0, 0) -- (0, 2);
        \draw [decoration={aspect=0.5, segment length=1.4mm, amplitude=.5mm, coil}, decorate] (0, 2) -- (1, 1.5);
        
        \draw [decoration={aspect=0.5, segment length=1.4mm, amplitude=.5mm, coil}, decorate] (1, 0) -- (2, .5);
        \draw [decoration={aspect=0.5, segment length=1.4mm, amplitude=.5mm, coil}, decorate] (1, 0) -- (0.25, 1.25);
        \draw [decoration={aspect=0.5, segment length=1.4mm, amplitude=.5mm, coil}, decorate] (0.25, 1.25) -- (2., 0.5);
        
        \draw [decoration={aspect=0.5, segment length=.8mm, amplitude=.5mm, coil}, decorate] (0, 0) -- (1., 0);
        \draw [decoration={aspect=0.5, segment length=.8mm, amplitude=.5mm, coil}, decorate] (0.25, 1.25) -- (1, 1.5);
        
        \draw [thick, fill = red!60] (0, 0) circle (.15);
        \draw [thick, fill = red!60] (0, 2) circle (.15);
        \draw [thick, fill = red!60] (1, 1.5) circle (.15);

        \draw [thick, fill = blue!60] (1, 0) circle (.15);
        \draw [thick, fill = blue!60] (0.25, 1.25) circle (.15);
        \draw [thick, fill = blue!60] (2, 0.5) circle (.15);
            
    \end{tikzpicture}
    }
    \caption{A spring network.}
    \label{fig:spring-network}
\end{figure}

It is evident that the objective function $E_p$ remains invariant under the group actions of $\mathrm{Iso}(E)$, or formally:
\begin{theorem}
\label{thm:invariant}
	For any $\sigma\in \mathrm{Iso}(E)$, any matrix $\bs{M}\in \mathbb{R}^{n\times n}$, and any embedding $\bs{Z}\in \mathbb{R}^{n\times d}$, we have $E_p(\sigma(\bs{Z}); \bs{M}, E)=E_p(\bs{Z}; \bs{M}, E)$.
\end{theorem}
The proof of this theorem can be found in Appendix \ref{app:prf-invariant}.

Several well-known spectral and spatial GNNs, such as GCN~\cite{kipf2017semi}, SGC~\cite{wu2019simplifying}, APPNP~\cite{klicpera2019predict}, GCNII~\cite{chen2020simple} and BernNet~\cite{he2021bernnet}, utilize the normalized Laplacian matrix $\tilde{\bs{L}}=\bs{D}^{-1/2}\bs{L}\bs{D}^{-1/2} = \bs{I} - \bs{D}^{-1/2}\bs{A}\bs{D}^{-1/2}$ instead of the standard Laplacian matrix $\bs{L} = \bs{D} - \bs{A}$ to enhance the numerical stability.
Following the convention established in related works, we also normalize our objective function by reparameterizing $\bs{Z}$ as $\bs{D}^{-1/2}\bs{Z}$, resulting in 
\begin{equation}
    \tilde{E}_p(\bs{Z};\bs{M}, E) = E_p(\bs{D}^{-1/2}\bs{Z};\bs{M}, E).
\end{equation} 
Furthermore, we add a trade-off regularization term to achieve the final optimization objective:
\begin{equation}
\label{eqn:objective}
    \mathcal{L}(\bs{Z}; \bs{Z}^{(0)}, \bs{M}, E) = (1-\alpha)\tilde{E}_p(\bs{Z}; \bs{M}, E) + \alpha \Vert\bs{Z} - \bs{Z}^{(0)}\Vert_F^2,
\end{equation}
where $\bs{Z}^{(0)} = f_\theta(\bs{X})$ is the initial embedding matrix of node features (see the framework of the model in Section \ref{subsec:solution}).

\subsection{Design of the Metric Matrix}
Another crucial aspect of our model is the design of the metric matrix.
In this section, we will discuss various ways to design it.

\paragraph{Pre-Designed Metric Matrix $\bs{M}$.}
One straightforward approach to obtain the elements in $\bs{M}$ is to utilize a pre-designed metric matrix.
For example, in certain scenarios like molecular conformation generation or graph drawing, if we possess prior knowledge about bond lengths or distances between nodes, we can directly assign the metric matrix for the MGNN model.
Theoretically, we can also set the metric matrix according to the output embeddings of any given GNN.
However, as it is hard to know the metric matrix in advance, this approach is only practical under some specific circumstances.

\paragraph{Learning Metric Matrix $\bs{M}$.}
A more general and reasonable approach is to learn the metric matrix from data.
Intuitively, to differentiate between nodes, we aim to increase the distance between dissimilar nodes and reduce the distance between similar nodes.
We can introduce a set of variables called edge attention, denoted as $\alpha_{ij}\in [-1, 1]$, associated with each edge to reflect the similarity between the two nodes connected by the edge.
Indeed, when the value of $\alpha_{ij}$ approaches $1$, it indicates that nodes $i$ and $j$ tend to belong to the same class. 
On the other hand, when $\alpha_{ij}$ approaches $-1$, it suggests that nodes $i$ and $j$ are more likely to belong to different classes.
The design of the edge attention is inspired by research on heterophilic graphs, where nodes with different labels tend to be connected by edges~\cite{pei2020geom, zhu2020beyond, chien2021adaptive, bo2021beyond, wang2021tree, he2021bernnet, yang2021graph, yang2021diverse}. 
Additionally, it draws inspiration from signed graphs, where edges are labeled with either +1 or -1, indicating positive or negative associations~\cite{harary1953on, derr2018signed, huang2019signed}. 
We suggest employing the following commonly used attention mechanisms to obtain the edge attention $\alpha_{ij}$:
\begin{itemize}
	\item ``concat'': $\alpha_{ij} = \tanh\left(\bs{a}^\top\left[\bs{H}_{i:}^\top\Vert\bs{H}_{j:}^\top\right]\right)$;
	\item ``bilinear'': $\alpha_{ij} = \tanh\left(\bs{H}_{i:}\bs{W}\bs{H}_{j:}^\top\right)$.
\end{itemize}
The matrix $\bs{H} = \mathrm{MLP}(\bs{Z}^{(0)})$ consists of hidden representations learned from the initial node embeddings.
Then we leverage these representations to learn the edge attention $\alpha_{ij}$.
Based on the meaning of $\alpha_{ij}$, we can design the elements $\bs{M}_{ij}$ in the metric matrix. Specifically, as $\alpha_{ij}\to 1$, we desire $\bs{M}_{ij}$ to approach $0$, and as $\alpha_{ij}\to -1$, we want $\bs{M}_{ij}$ to tend towards $+\infty$.
One simple design that fulfills these conditions is to set $\bs{M}_{ij}$ as follows:
\begin{equation}
    \bs{M}_{ij} = \frac{1 - \alpha_{ij}}{1 + \alpha_{ij} + \varepsilon}\Vert\bs{Z}^{(0)}_{i:} - \bs{Z}^{(0)}_{j:}\Vert_2, 
\end{equation}
where $\varepsilon$ is a small positive number, and $\bs{Z}^{(0)}$ is the initial embedding matrix.

\subsection{Solution and Framework}
\label{subsec:solution}
In this section, we will present the complete framework of the model.

\paragraph{Embedding.} 
Following the standard scheme of GNNs, we first embed the raw node features into a $d$-dimensional latent space via an embedding function $\bs{Z}^{(0)} = f_\theta(\bs{X})$ to reduce the dimension. 
A linear layer 
\begin{equation}
    f_{\bs{W}, \bs{b}}(\bs{X})=\bs{XW}+\bs{1}\bs{b}^\top
\end{equation} 
in linear GNNs or a two-layer MLP 
\begin{equation}
    f_{\bs{W}_1, \bs{W}_2, \bs{b}_1, \bs{b}_2}(\bs{X})=\sigma(\sigma(\bs{X}\bs{W}_1+\bs{1}\bs{b}_1^\top)\bs{W}_2 + \bs{1}\bs{b}_2^\top)
\end{equation}
in spectral GNNs are all good choices for the embedding function.

\paragraph{Propagation.}
In this section, for formula conciseness, we use the notation $\tilde{\bs{X}}$ to represent $\bs{D}^{-1/2}\bs{X}\bs{D}^{-1/2}$ for any matrix $\bs{X}$.

We will design a graph propagation method which tries to minimize the objective function in Equation \ref{eqn:objective}.
Unfortunately, the objective function is generally non-convex when for some $i, j$, $\bs{M}_{ij} > \Vert \bs{Z}_{i:} - \bs{Z}_{j:}\Vert_2$, which means it may have multiple local minima, making it challenging to find the global minimum.
Consequently, we can employ iterative methods such as gradient descent~\cite{kruskal1964multi, kruskal1964nonmetric} or stationary point iteration~\cite{guttman1968general} introduced in~\cite{de2005applications} to optimize the objective function $E_p(\bs{Z}; \bs{M}, E)$.
The gradient descent method leverages gradient information to iteratively update the embeddings until reaching a point with a near-zero gradient.
This is quite similar to the idea of stationary point iteration method which uses the first order necessary condition $\nabla f = \bs{0}$ to derive an iteration equation.
By employing the stationary point iteration method, we can derive a propagation equation that exhibits a similar form and maintains close connections to related works in the field.
To begin, let us compute the gradient of the final optimization objective function with respect to the embedding $\bs{Z}$:
\begin{equation}
    \label{eqn:gradient}
	\frac{\partial \mathcal{L}}{\partial \bs{Z}} = 2(1-\alpha)\bs{D}^{-1/2}(\mathrm{diag}(\bs{S}\bs{1}) - \bs{S})\bs{D}^{-1/2}\bs{Z} + 2\alpha(\bs{Z}-\bs{Z}^{(0)}),
\end{equation}
where $\bs{S} = \bs{A} - \bs{A}\odot\bs{M}\odot M(\bs{D}^{-1/2}\bs{Z})^{\odot -1}$.
Setting the gradient of $\mathcal{L}$ to zero and rearranging the terms, we obtain the following equation:
\begin{equation}
    \bs{Z} = (1 - \alpha)\tilde{\bs{A}}\bs{Z} + (1 - \alpha)\widetilde{\bs{L}_{\bs{H}}}\bs{Z} + \alpha \bs{Z}^{(0)},
\end{equation}
where $\bs{H} = \bs{A}\odot\bs{M}\odot M(\bs{D}^{-1/2}\bs{Z})^{\odot -1}$, and $\bs{L}_{\bs{H}} = \mathrm{diag}(\bs{H}\bs{1}) - \bs{H}$.
The meaning of using the notation $\bs{L}_{\bs{H}}$ is that if we have a graph $G$ whose adjacency matrix is $\bs{H}$, the Laplacian of $G$ is $\bs{L}_{\bs{H}}$.
We rewrite the equation as an iteration form, and substitute $1-\alpha$ with $\beta$ in the second term to allow more flexibility.
This leads us to the final propagation equation:
\begin{equation}
\label{eqn:propagate}
    \bs{Z}^{(k + 1)} = (1 - \alpha)\tilde{\bs{A}}\bs{Z}^{(k)} + \beta\widetilde{\bs{L}_{\bs{H}}}\bs{Z}^{(k)} + \alpha \bs{Z}^{(0)},
\end{equation}
We can also derive the message-passing form of the propagation rule as follows:
\begin{equation}
    \resizebox{1\linewidth}{!}{
    $\bs{Z}_{i:}^{(k + 1)} = (1-\alpha)\sum\limits_{j \in N(i)}\dfrac{\bs{Z}_{i:}^{(k)}}{\sqrt{d_id_j}} + \beta\sum\limits_{j\in \mathcal{N}(i)} \frac{\bs{M}_{ij}\left(\bs{Z}_{i:}^{(k)} - \bs{Z}_{j:}^{(k)}\right)}{\sqrt{d_id_j}\left\Vert\frac{\bs{Z}_{i:}^{(k)}}{\sqrt{d_i}} - \frac{\bs{Z}_{j:}^{(k)}}{\sqrt{d_j}}\right\Vert_2} + \alpha \bs{Z}_{i:}^{(0)}.$
    }
\end{equation}
We may refer to the first term, $\tilde{\bs{A}}\bs{Z}^{(k)}$, in Equation \ref{eqn:propagate} ``topological message'' since it aggregates the hidden features based on the graph topology.
Similarly, we can name the second term, $\widetilde{\bs{L}_{\bs{H}}}\bs{Z}^{(k)}$, the ``metric message'' as it aggregates the hidden features according to the metric matrix.
It is worth noting that this equation can be viewed as a generalization of the propagation equation in the APPNP model \cite{klicpera2019predict}.

\paragraph{Optional Linear and Non-Linear Transformations.}
For applications where a pre-designed metric matrix is not available, such as node classification, we can incorporate linear and non-linear transformations into our MGNN model, similar to the approach used in the GCNII model \cite{chen2020simple}. 
Now we have the updating rule for tasks which need to learn metric matrix from data:
\begin{equation}
\label{eqn:MGNN-update}
    \resizebox{\linewidth}{!}{
    $\bs{Z}^{(k + 1)} = \sigma(((1 - \alpha)\tilde{\bs{A}}\bs{Z}^{(k)} + \beta\widetilde{\bs{L}_{\bs{H}}}\bs{Z}^{(k)} + \alpha \bs{Z}^{(0)})({\gamma}^{(k)}\bs{W}^{(k)} + (1-{\gamma}^{(k)})\bs{I})),$
    }
\end{equation}
where $\gamma^{(k)}$ is set as in the GCNII model. 
That is, $\gamma^{(k)} = \log(1 + \theta/k)$, and $\theta$ is a hyper-parameter.
For applications where a pre-designed metric matrix is available in advance, such as graph drawing, we do not include additional linear and non-linear transformations in the propagation rule.

\paragraph{Classification.} 
Finally, we map the hidden representation vector to the output dimension using a classification function $\bs{Y} = g_\theta(\bs{Z}^{(L)})$.
We use a single-layer linear model 
\begin{equation}
g_{\bs{W}, \bs{b}}(\bs{Z}^{(L)}) = \bs{Z}^{(L)}\bs{W}+\bs{1}\bs{b}^\top
\end{equation}
as the final classification function.

\section{Experiments}
In this section, we conduct experiments on both synthetic and real-world graphs to evaluate the performance of our MGNN model.
All experiments are conducted on a server with an Intel Xeon Silver 4114 CPU (2.20GHz), an Nvidia Quadro RTX 8000 GPU (48GB) and 1TB of RAM.

\subsection{Arranging Nodes with the Given Metric Matrices}
\label{subsec:arrange}
To verify that our MGNN model can arrange the nodes with the given metric matrix correctly, we perform experiments on synthetic stochastic block model graphs.

\paragraph{Datasets.}
We randomly generate two stochastic block model (SBM) graphs: one homophilic and one heterophilic.
Each graph comprises 4 blocks, with 50 nodes in each block.
In the homophilic graph, the probability of intra-block edges is 0.3, while the probability of inter-class edges is 0.05.
In the heterophilic graph, the probabilities are 0.05 and 0.2 for intra-block and inter-class edges, respectively.
The node features in each block are sampled from four 2-dimensional Gaussian distributions, with parameters $(\bs{\mu} = [0, 0]^\top, \bs{\Sigma} = \bs{I})$, $(\bs{\mu} = [1, 0]^\top, \bs{\Sigma} = \bs{I})$, $(\bs{\mu} = [0, 1]^\top, \bs{\Sigma} = \bs{I})$, $(\bs{\mu} = [1, 1]^\top, \bs{\Sigma} = \bs{I})$.

\begin{figure}[ht]
    \centering
    \includegraphics[width = 0.4\textwidth]{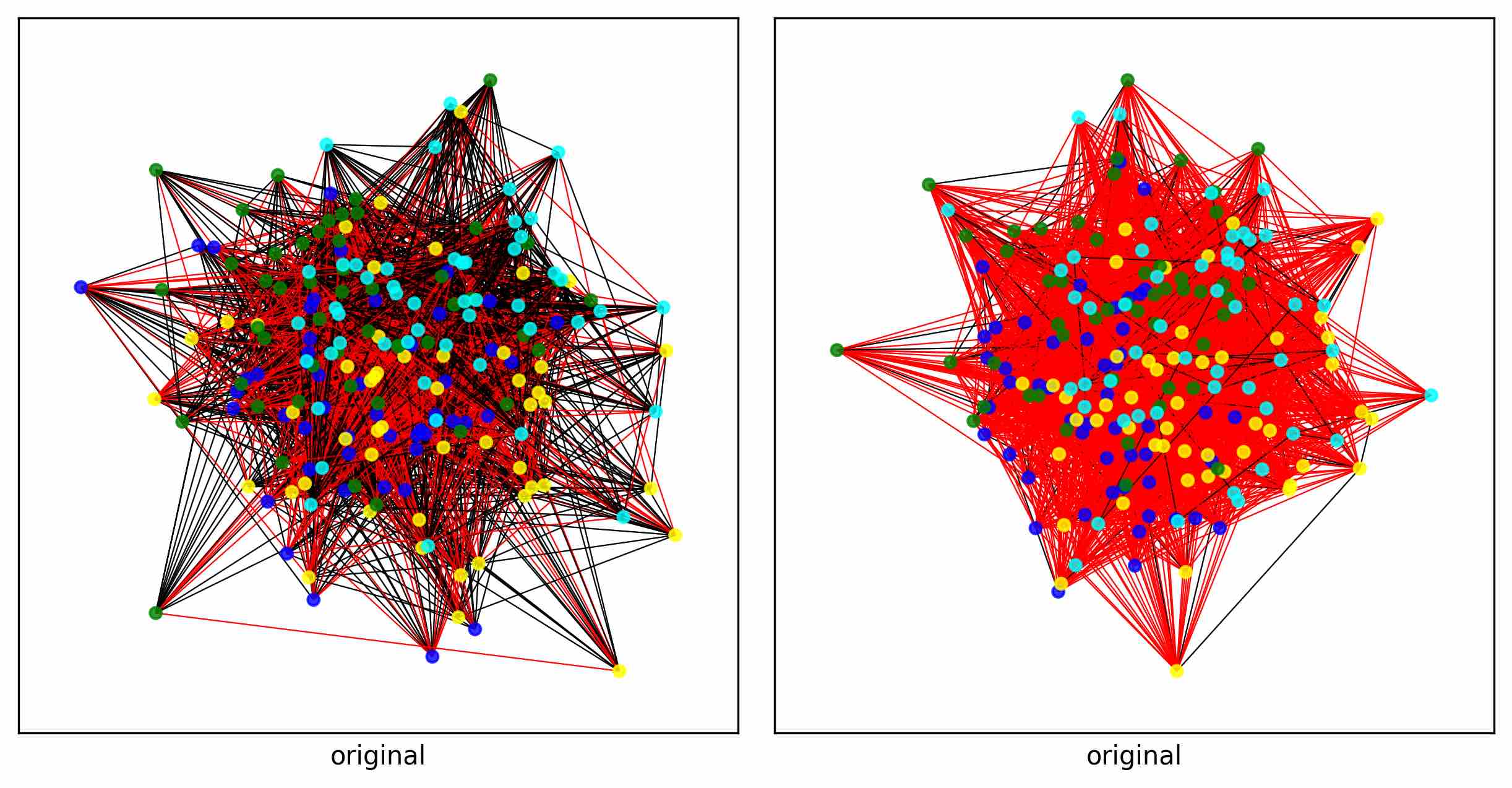}
    \caption{Generated homophilic (left) and heterophilic (right) graphs. Black edges are intra-class, while red edges are inter-class.}
    \label{fig:original}
\end{figure}

\begin{figure*}[ht]
    \centering
    \includegraphics[width = \textwidth]{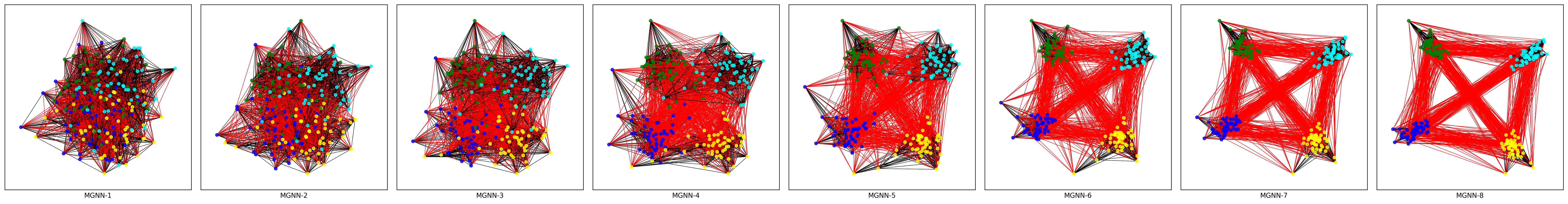}
    \caption{The results of the homophilic SBM graph of MGNN layers.}
    \label{fig:MGNN-homo}
\end{figure*}

\begin{figure*}[ht]
    \centering
    \includegraphics[width = \textwidth]{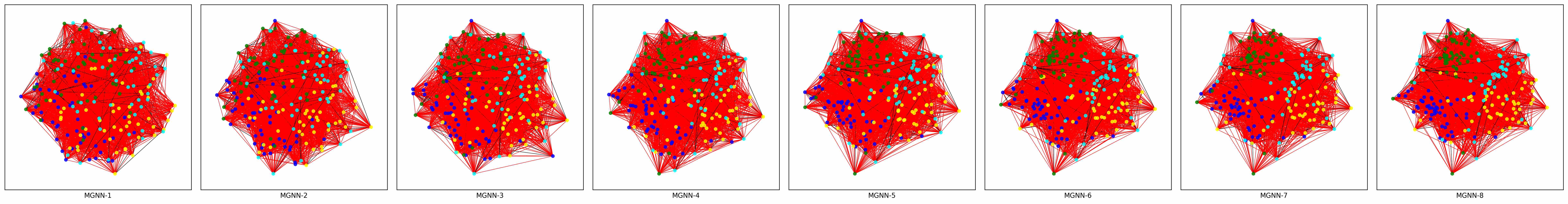}
    \caption{The results of the heterophilic SBM graph of MGNN layers.}
    \label{fig:MGNN-hetero}
\end{figure*}

\paragraph{Settings and Results.}
We first generate two graphs as described above and visualize them in Figure \ref{fig:original}.
Next, we assign metric matrices $\bs{M}$ to the two graphs. 
Specifically, for nodes $i$ and $j$ belonging to the same block, $\bs{M}_{ij} = 0$, while for nodes in different blocks, $\bs{M}_{ij} = 5$.
We propagate the node features through 8 MGNN propagation layers, using parameters $\alpha=0.05$ and $\beta=0.5$. 
After each round of propagation, we visualize the results.
The results are visualized in Figure \ref{fig:MGNN-homo} and \ref{fig:MGNN-hetero}.
Based on the results, we can confirm that our MGNN model aims to separate the four blocks.
Moreover, for the homophilic graph, our MGNN model can arrange the nodes with the given metric matrix correctly.
This is because the presence of homophilic edges adds convexity to our objective function, whereas heterophilic edges have the opposite effect, and it is known that convex functions are easier to optimize.

\paragraph{Compare MGNN with SGC and APPNP Layers.}
Additionally, we provide visualizations of the results obtained from the SGC layers (refer to Figure \ref{fig:SGC-homo} and Figure \ref{fig:SGC-hetero} in Appendix \ref{app:vis-results}) and APPNP layers (refer to Figure \ref{fig:APPNP-homo} and Figure \ref{fig:APPNP-hetero} in Appendix \ref{app:vis-results}) for the two synthetic graphs mentioned earlier.
The results indicate that as we stack SGC layers, 
the embeddings of the nodes tend to converge towards a linear subspace, which will make the embedding vectors difficult to classify.
Furthermore, we can conclude that while stacking APPNP layers does not exhibit the same phenomenon as described above, the APPNP model is unable to arrange the nodes in accordance with the metric matrix as the MGNN layers.

\subsection{Supervised Node Classification}
To assess the real-world performance of our MGNN model, we conduct node-classification experiments and compare it with other models.
\paragraph{Datasets.}
As for the datasets, we carefully select many well-known benchmark graphs including the citation networks Cora, CiteSeer, PubMed from~\cite{yang2016revisiting}; the CoraFull dataset from~\cite{bojchevski2018deep}; the coauthor networks CS and Physics from~\cite{shchur2018pitfalls}; the WebKB networks Cornell, Texas and Wisconsin from~\cite{pei2020geom}; the Wikipedia networks Chameleon and Squirrel from~\cite{rozembercaki2021multi}; the actor co-occurrence network Actor from~\cite{pei2020geom}; the WikiCS dataset from~\cite{mernyei2020wiki}; and the ogbn-arxiv dataset from~\cite{hu2020open}.
We download and preprocess all datasets using either the PyTorch Geometric Library~\cite{fey2019fast} or the official OGB package~\cite{hu2020open}.
The basic statistics of the datasets, including the number of nodes, number of edges, number of features, number of classes, and the edge homophily (as defined in ~\cite{zhu2020beyond}) denoted as $\mathcal{H}(G, Y)$, are summarized in Table \ref{tab:statistics}.
The datasets encompass a wide range, from small graphs with approximately 200 nodes and 600 edges, to large graphs with around 170,000 nodes and over 2 million edges. 
Among these datasets, there are highly homophilic graphs with edge homophily values exceeding 0.9, as well as heterophilic graphs exhibiting homophily values below 0.1.

\begin{table}[ht]
    \centering
    \caption{The statistics of the datasets.}
    \label{tab:statistics}
    \resizebox{\linewidth}{!}{
    \begin{threeparttable}
    \begin{tabular}{cccccc}
        \toprule
        Dataset         & \#Nodes       & \#Edges\tnote{1}  & \#Features        & \#Classes     & $\mathcal{H}(G, Y)$ \\
        \midrule
        Cora            & 2,708         & 10,556            & 1,433             & 7             & 0.81 \\
        CiteSeer        & 3,327         & 9,104             & 3,703             & 6             & 0.74 \\
        PubMed          & 19,717        & 88,648            & 500               & 3             & 0.80 \\
        CoraFull        & 19,793        & 126,842           & 8,710             & 70            & 0.57 \\
        CS              & 18,333        & 163,788           & 6,805             & 15            & 0.81 \\
        Physics         & 34,493        & 495,924           & 8,415             & 5             & 0.93 \\
        Cornell         & 183           & 557               & 1,703             & 5             & 0.13 \\
        Texas           & 183           & 574               & 1,703             & 5             & 0.09 \\
        Wisconsin       & 251           & 916               & 1,703             & 5             & 0.19 \\
        Chameleon       & 2,277         & 62,792            & 2,325             & 5             & 0.23 \\
        Squirrel        & 5,201         & 396,846           & 2,089             & 5             & 0.22 \\
        Actor           & 7,600         & 53,411            & 932               & 5             & 0.22 \\
        WikiCS          & 11,701        & 431,726           & 300               & 10            & 0.65 \\
        ogbn-arxiv      & 169,343       & 2,315,598         & 128               & 40            & 0.65 \\
        \bottomrule
    \end{tabular}
    \begin{scriptsize}
    \begin{tablenotes}
        \item[1] We first transform directed graphs to undirected ones, then count the total number of pairs $(i, j)$ in the edge sets for each dataset.
    \end{tablenotes}
    \end{scriptsize}
    \end{threeparttable}
    }
\end{table}

\paragraph{Methods.}
To demonstrate the effectiveness of the design of our MGNN model, we have selected various baseline models, including non-graph models, spectral GNNs, and spatial (non-spectral) GNN models.
For the {\bfseries non-graph models}, we have chosen the one-layer linear layer model (with bias) and the two-layer MLP model as our baselines. 
For the {\bfseries spectral GNNs}, we have chosen the GCN model~\cite{kipf2017semi}, the SGC model~\cite{wu2019simplifying}, and the APPNP model~\cite{klicpera2019predict}. 
For the {\bfseries spatial (non-spectral) GNNs}, we have selected the PointNet model~\cite{qi2017pointnet}, the GAT model~\cite{velickovic2018graph}, the GIN model~\cite{xu2019how}, the GCNII model ~\cite{chen2020simple}, the $^p$GNN model~\cite{fu2022p}, and the LINKX model~\cite{lim2021large}. 

\paragraph{Efficiency of the Models.}
In this part, we will analyze the total number of trainable parameters and the time complexity of the forward pass for each model.
We denote $F$ as the number of features, representing the input dimension of the dataset. Similarly, $c$ represents the number of classes, which indicates the output dimension of the dataset. We will use $d$ to represent the hidden dimension, and $L$ to denote the number of convolution layers or propagation depth.
Meanwhile, suppose that a graph $G$ has $n=\vert V\vert$ nodes and has $m=\vert E\vert$ edges.

For the one-layer linear model ($\bs{Y} = \bs{X}\bs{W}_{F\times c} + \bs{1}\bs{b}_{c}^\top$), it has $O(Fc)$ trainable parameters, and its forward pass has a time complexity of $O(nFc)$.
Regarding the $L$-layer MLP, each layer operates similarly to a linear layer, with the addition of a non-linear activation function. 
Consequently, each MLP layer possesses $O(\text{dim\_in}\times \text{dim\_out})$ trainable parameters, and the forward pass requires a time complexity of $O(n\times\text{dim\_in}\times \text{dim\_out})$.

The cases of the GNNs are complicated, so we first divide them into three phases and then analyze them individually.
\begin{itemize}
    \item Embedding Phase: 
        In the GCN, GAT and the GIN model, the embedding layers serve as their initial layers for dimension reduction. 
        Consequently, they have $O(Fd)$ parameters ($O(Fd + d^2) = O(Fd)$ for GIN, as it utilizes a two-layer MLP as the update function), and their forward pass has a time complexity of $O(mF + nFd)$ (the $O(mF)$ portion accounts for the graph propagation operation, as explained below).
        In the case of the MGNN$_{\text{Linear}}$ model, the embedding function is implemented as a linear layer, which contains $O(Fd)$ trainable parameters and has a forward time complexity of $O(nFd)$.
        For the PointNet model, the APPNP model, the GCNII model, and the MGNN$_{\text{MLP}}$ model, their embedding function consists of a two-layer MLP. 
        This MLP comprises $O(Fd + d^2) = O(Fd)$ trainable parameters and has a forward time complexity of $O(nFd + nd^2) = O(nFd)$.
        Regarding the $^p$GNN model, its embedding function is a one-layer MLP, which possesses $O(Fd)$ parameters and has a forward time complexity of $O(nFd)$.

    \item Propagation / Convolution Phase:
        For message-passing GNNs (GNNs excluding the LINKX model), their propagation rules are $\bs{Z}_{i:}^{(k + 1)} = \mathrm{UPDATE}(\bs{Z}_{i:}^{(k)}, \mathrm{AGG}_{ j\in N(i)} \mathrm{MSG}(\bs{Z}_{i:}^{(k)}, \bs{Z}_{j:}^{(k)}))$.\\
        These rules do not involve any trainable parameters and have a time complexity of $\sum_{i\in V}d_id = O(md)$, assuming that $\mathrm{MSG}(\bs{Z}_{i:}^{(k)}, \bs{Z}_{j:}^{(k)})$ takes $O(d)$ time.
        For GCN, GAT, GIN, GCNII, PointNet, and our MGNN models, their convolution modules incorporate linear transformations, resulting in $O(d^2)$ parameters and an overall time complexity of $O(nd^2)$ per layer.
        In the case of GAT, each layer introduces an additional $O(d)$ parameters to learn the attention.

    \item Classification Phase: 
        For the GCN, GAT and the GIN model, the classification layers are their last layers.
        They possess $O(dc)$ ($O(d^2 + dc)$ for GIN) parameters, and their forward time complexity is $O(md + ndc)$ ($O(md + nd^2 + ndc)$ for GIN).
        For other message-passing GNNs, their classification function is implemented as a linear layer, which has $O(dc)$ ($O(Fc)$ for SGC) parameters and $O(ndc)$ ($O(nFc)$ for SGC) forward time complexity.

    \item Other Modules: The LINKX model is not a normal message-passing GNN. Its structure is as follows:
    \begin{align}
    \begin{split}
        &\bs{H}_{\bs{A}} = \mathrm{MLP}_{\bs{A}}(\bs{A}), \bs{H}_{\bs{X}} = \mathrm{MLP}_{\bs{X}}(\bs{X}), \\
        &\bs{H} = \sigma([\bs{H}_{\bs{A}}\Vert\bs{H}_{\bs{X}}]\bs{W} + \bs{H}_{\bs{A}} + \bs{H}_{\bs{X}}), \\
        &\bs{Y} = \mathrm{MLP}_{f}(\bs{H}). 
    \end{split}
    \end{align}
    We denote $L_{\bs{A}}$ as the number of layers in $\mathrm{MLP}_{\bs{A}}$, $L_{\bs{X}}$ as the number of layers in $\mathrm{MLP}_{\bs{X}}$, $L_f$ as the number of layers in $\mathrm{MLP}_f$, and $L = L_{\bs{A}} + L_{\bs{X}} + L_f$ represents the total number of layers.
    The $\mathrm{MLP}_{\bs{A}}$ has $O(nd + d^2L_{\bs{A}})$ parameters and the time complexity is $O(md + nd^2L_{\bs{A}})$. 
    Note that the first layer, $\bs{H}_{\bs{A}}^{(1)} = \sigma(\bs{A}\bs{W}_0 + \bs{1}\bs{b}_0^\top)$, can be implemented with sparse matrix multiplication, resulting in a time complexity of $O(md)$.
    Similarly, the $\mathrm{MLP}_{\bs{X}}$ has $O(Fd + d^2L_{\bs{X}})$ parameters and the time complexity is $O(nFd + nd^2L_{\bs{X}})$.
    The computation of $\bs{H}$ requires $O(d^2)$ parameters and takes $O(nd^2)$ time.
    Lastly, the $\mathrm{MLP}_f$ has $O(d^2L_f + dc)$ trainable parameters, and its forward time complexity is $O(nd^2L_f + ndc)$.
    In the case of the PointNet model, there is a position generation layer that comprises a two-layer MLP. 
    This layer possesses $O(d^2)$ trainable parameters, and the forward time complexity is $O(nd^2)$.
    As for our MGNN model, it incorporates a metric layer consisting of a two-layer MLP and an attention module.
    The metric layer has $O(d^2)$ trainable parameters, and its time complexity is $O(md^2)$.
\end{itemize}

We summarize the total parameters of each model in Table \ref{tab:parameters}, and their forward pass time complexity in Table \ref{tab:time}.

\begin{table}[ht]
    \centering
    \caption{The total number of trainable parameters of each model.}
    \label{tab:parameters}
    \resizebox{\linewidth}{!}{
    \begin{tabular}{c|cccc|c}
        \toprule
         & Embedding & Prop./Conv./Hidden & Classification & Others & Total \\ 
        \midrule
        Linear & --- & --- & $O(Fc)$ & --- & $O(Fc)$ \\
        MLP & $O(Fd)$ & $O(d^2L)$ & $O(dc)$ & --- & $O(d(F + dL + c))$ \\
        GCN & $O(Fd)$ & $O(d^2L)$ & $O(dc)$ & --- & $O(d(F + dL + c))$ \\
        SGC & $0$ & $0$ & $O(Fc)$ & --- & $O(Fc)$ \\
        APPNP & $O(Fd)$ & $0$ & $O(dc)$ & --- & $O(d(F + d + c))$ \\
        PointNet & $O(Fd)$ & $O(d^2L)$ & $O(dc)$ & $O(d^2)$ & $O(d(F + dL + c))$ \\
        GAT & $O(Fd)$ & $O(d^2L)$ & $O(dc)$ & --- & $O(d(F + dL + c))$ \\
        GIN & $O(Fd)$ & $O(d^2L)$ & $O(d^2 + dc)$ & --- & $O(d(F + dL + c))$ \\
        GCNII & $O(Fd)$ & $O(d^2L)$ & $O(dc)$ & --- & $O(d(F + dL + c))$ \\
        $^p$GNN & $O(Fd)$ & $0$ & $O(dc)$ & --- & $O(d(F+c))$ \\
        \midrule
        MGNN$_{\text{Linear}}$ & \multirow{2}{*}{$O(Fd)$} & \multirow{2}{*}{$O(d^2L)$} & \multirow{2}{*}{$O(dc)$} & \multirow{2}{*}{$O(d^2)$} & \multirow{2}{*}{$O(d(F + dL + c))$} \\
        MGNN$_{\text{MLP}}$ & & & & & \\
        \bottomrule
    \end{tabular}
    }
    
    \resizebox{\linewidth}{!}{
    \begin{tabular}{c|cccc|c}
        \toprule
         & $\mathrm{MLP}_{\bs{A}}$ & $\mathrm{MLP}_{\bs{X}}$ & Computing $\bs{H}$ & $\mathrm{MLP}_f$ & Total \\ 
        \midrule
        LINKX & $O(nd+ d^2L_{\bs{A}})$ & $O(Fd + d^2L_{\bs{X}})$ & $O(d^2)$ & $O(d^2L_f + dc)$ & $O(d(dL + n + F + c))$ \\
        \bottomrule
    \end{tabular}
    }
\end{table}

\begin{table}[ht]
    \centering
    \caption{The time complexity of each model.}
    \label{tab:time}
    \resizebox{\linewidth}{!}{
    \begin{tabular}{c|cccc|c}
        \toprule
         & Embedding & Prop./Conv./Hidden & Classification & Others & Total \\ 
        \midrule
        Linear & --- & --- & $O(nFc)$ & --- & $O(nFc)$ \\
        MLP & $O(nFd)$ & $O(nd^2L)$ & $O(ndc)$ & --- & $O(nd(F + dL + c))$ \\
        GCN & $O(mF + nFd)$ & $O((md + nd^2)L)$ & $O(md + ndc)$ & --- & $O(md(F + dL + c))$ \\
        SGC & --- & $O(mFL)$ & $O(nFc)$ & --- & $O(mF(L + c))$ \\
        APPNP & $O(nFd)$ & $O(mdL)$ & $O(ndc)$ & --- & $O(nd(F + d + c) + mdL)$ \\
        PointNet & $O(mF + nFd)$ & $O((md + nd^2)L)$ & $O(md + ndc)$ & $O(nd^2)$ & $O(md(F + dL + c))$ \\
        GAT & $O(mF + nFd)$ & $O((md + nd^2)L)$ & $O(md + ndc)$ & --- & $O(md(F + dL + c))$ \\
        GIN & $O(mF + nFd)$ & $O((md + nd^2)L)$ & $O(md + nd^2 + ndc)$ & --- & $O(md(F + dL + c))$ \\
        GCNII & $O(nFd)$ & $O((md + nd^2)L)$ & $O(ndc)$ & --- & $O(nd(F + d + c) + md^2L)$ \\
        $^p$GNN & $O(nFd)$ & $O(mdL)$ & $O(ndc)$ & --- & $O(nd(F + c) + mdL)$ \\
        \midrule
        MGNN$_{\text{Linear}}$ & \multirow{2}{*}{$O(nFd)$} & \multirow{2}{*}{$O((md + nd^2)L)$} & \multirow{2}{*}{$O(ndc)$} & \multirow{2}{*}{$O(md^2)$} & \multirow{2}{*}{ $O(nd(F + d + c) + md^2L)$}\\ 
        MGNN$_{\text{MLP}}$ & & & & & \\
        \bottomrule
    \end{tabular}
    }
    \resizebox{\linewidth}{!}{
    \begin{tabular}{c|cccc|c}
        \toprule
         & $\mathrm{MLP}_{\bs{A}}$ & $\mathrm{MLP}_{\bs{X}}$ & Computing $\bs{H}$ & $\mathrm{MLP}_f$ & Total \\ 
        \midrule
        LINKX & $O(md+ nd^2L_{\bs{A}})$ & $O(nFd + nd^2L_{\bs{X}})$ & $O(nd^2)$ & $O(nd^2L_f + ndc)$ & $O(d(ndL + m + nF + nc))$\\
        \bottomrule
    \end{tabular}
    }
\end{table}

\begin{table*}[ht]
	\centering
	\caption{The results (accuracy with standard deviation) of the supervised node classification experiments. Boldface results indicate the best model on each dataset, and \ul{underlined} results are second best models.}
	\label{tab:supervised}
	\resizebox{\textwidth}{!}{
	\begin{tabular}{c||cc|ccc|cccccc||c}
		\toprule
                        & \multicolumn{2}{c|}{\textbf{Non-Graph}}                   & \multicolumn{3}{c|}{\textbf{Spectral}}                                        & \multicolumn{6}{c||}{\textbf{Spatial (Non-Spectral)}}                                                          & \multirow{2}{*}{MGNN}     \\
						& Linear			       & MLP 				        & GCN 				        & SGC                        & APPNP 			        & PointNet                  & GAT 				        & GIN                       & GCNII				        & $^p$GNN                       & LINKX                     & \\
		\midrule
		Cora 			& $76.09_{\pm 1.55}$       & $76.11_{\pm 1.58}$ 	    & $88.25_{\pm 1.09}$	    & $88.34_{\pm 1.41}$         & $\bs{89.30_{\pm 1.45}}$	& $84.43_{\pm 1.94}$        & $88.71_{\pm 0.89}$	    & $85.76_{\pm 1.18}$        & $88.52_{\pm 1.40}$	    & $88.78_{\pm 1.20}$         & $82.82_{\pm 1.96}$        & $\ul{89.04_{\pm 1.20}}$	    \\
		CiteSeer		& $71.11_{\pm 1.81}$	   & $73.25_{\pm 1.16}$ 	    & $77.10_{\pm 1.12}$	    & $\ul{77.47_{\pm 1.42}}$    & $76.80_{\pm 1.10}$	    & $72.83_{\pm 1.38}$        & $76.44_{\pm 1.19}$	    & $72.83_{\pm 1.47}$        & $77.05_{\pm 0.78}$	    & $77.28_{\pm 0.88}$         & $71.79_{\pm 1.55}$        & $\bs{78.08_{\pm 1.50}}$	\\
		PubMed			& $87.06_{\pm 0.67}$	   & $88.21_{\pm 0.46}$ 	    & $89.07_{\pm 0.42}$	    & $87.59_{\pm 0.54}$         & $89.75_{\pm 0.47}$	    & $89.18_{\pm 0.38}$        & $87.77_{\pm 0.67}$	    & $87.15_{\pm 0.54}$        & $\ul{90.24_{\pm 0.55}}$	& $89.79_{\pm 0.38}$         & $86.77_{\pm 0.59}$        & $\bs{90.37_{\pm 0.47}}$     \\    
		CoraFull    	& $60.55_{\pm 0.76}$	   & $61.61_{\pm 0.56}$	        & $71.86_{\pm 0.75}$	    & $71.86_{\pm 0.70}$         & ${71.88_{\pm 0.75}}$	    & $63.32_{\pm 1.02}$        & $70.65_{\pm 0.86}$	    & $67.11_{\pm 0.46}$        & $71.09_{\pm 0.72}$	    & $\ul{72.36_{\pm 0.58}}$        & $63.57_{\pm 0.55}$        & $\bs{72.42_{\pm 0.69}}$	\\
		CS           	& $94.51_{\pm 0.32}$	   & $94.89_{\pm 0.23}$	        & $93.77_{\pm 0.37}$	    & $94.10_{\pm 0.40}$         & $95.91_{\pm 0.23}$	    & $93.13_{\pm 0.42}$        & $93.25_{\pm 0.31}$	    & $91.81_{\pm 0.34}$        & $\ul{95.98_{\pm 0.22}}$	& $95.83_{\pm 0.23}$         & $95.06_{\pm 0.24}$        & $\bs{96.01_{\pm 0.16}}$	\\
		Physics      	& $95.92_{\pm 0.15}$	   & $96.01_{\pm 0.26}$	        & $96.46_{\pm 0.25}$	    & OOM                        & $\ul{97.14_{\pm 0.21}}$	& $96.37_{\pm 0.27}$        & $96.47_{\pm 0.21}$	    & $94.66_{\pm 2.04}$        & $97.10_{\pm 0.21}$	    & $96.93_{\pm 0.10}$         & $96.87_{\pm 0.16}$        & $\bs{97.15_{\pm 0.15}}$\\
		Cornell			& $\ul{78.65_{\pm 2.82}}$  & $75.95_{\pm 4.43}$	        & $50.27_{\pm 6.86}$	    & $53.51_{\pm 4.32}$         & $77.84_{\pm 6.49}$	    & $71.08_{\pm 5.93}$        & $50.27_{\pm 7.66}$	    & $49.46_{\pm 7.46}$        & $76.76_{\pm 7.76}$	    & $77.30_{\pm 8.22}$         & $71.08_{\pm 12.09}$       & $\bs{81.89_{\pm 6.29}}$  \\
		Texas			& $81.35_{\pm 4.90}$	   & $83.78_{\pm 7.55}$	        & $61.08_{\pm 8.65}$	    & $56.22_{\pm 6.37}$         & $86.76_{\pm 1.46}$	    & $82.16_{\pm 6.30}$        & $61.08_{\pm 5.57}$	    & $64.05_{\pm 4.20}$        & $\ul{88.65_{\pm 4.95}}$	& $85.14_{\pm 5.16}$         & $84.32_{\pm 5.24}$        & $\bs{90.00_{\pm 2.72}}$	\\
		Wisconsin		& $86.20_{\pm 3.63}$	   & $\bs{88.80_{\pm 2.40}}$	& $55.80_{\pm 5.83}$	    & $58.00_{\pm 4.29}$         & $86.00_{\pm 3.10}$	    & $81.60_{\pm 3.98}$        & $56.40_{\pm 5.99}$	    & $57.20_{\pm 6.82}$        & $88.20_{\pm 4.14}$	    & $85.40_{\pm 3.58}$         & $82.00_{\pm 3.10}$        & $\ul{88.40_{\pm 3.44}}$		\\
		Chameleon		& $50.77_{\pm 2.07}$	   & $50.75_{\pm 2.13}$	        & $\ul{69.21_{\pm 2.08}}$	& $67.12_{\pm 2.17}$         & $67.85_{\pm 2.65}$	    & $63.76_{\pm 2.52}$        & $66.97_{\pm 2.45}$	    & $46.73_{\pm 13.51}$       & $67.56_{\pm 1.18}$	    & $69.19_{\pm 1.57}$         & $67.47_{\pm 1.62}$        & $\bs{72.37_{\pm 2.25}}$	\\
		Squirrel		& $35.76_{\pm 1.05}$	   & $35.79_{\pm 1.65}$	        & $55.43_{\pm 2.05}$	    & $52.18_{\pm 1.49}$         & $54.60_{\pm 1.88}$	    & $47.39_{\pm 8.31}$        & $\ul{55.68_{\pm 2.81}}$	& $20.96_{\pm 2.08}$        & $53.88_{\pm 2.77}$	    & $51.61_{\pm 1.28}$         & $\bs{57.86_{\pm 1.17}}$   & $54.45_{\pm 1.85}$		\\
		Actor			& $36.16_{\pm 0.75}$	   & $37.67_{\pm 1.60}$	        & $30.42_{\pm 1.58}$	    & $30.14_{\pm 1.18}$         & $36.98_{\pm 1.28}$	    & $36.41_{\pm 1.23}$        & $29.22_{\pm 0.94}$	    & $26.09_{\pm 1.75}$        & $\ul{37.75_{\pm 1.23}}$	& $36.47_{\pm 1.07}$         & $35.00_{\pm 2.11}$        & $\bs{38.36_{\pm 1.46}}$	\\
		WikiCS          & $78.74_{\pm 0.57}$       & $79.85_{\pm 0.69}$         & $84.02_{\pm 0.61}$        & $83.47_{\pm 0.83}$         & $84.88_{\pm 0.55}$       & $84.09_{\pm 0.86}$        & $83.82_{\pm 0.73}$        & $66.08_{\pm 22.77}$       & $\ul{85.09_{\pm 0.71}}$   & $84.41_{\pm 0.46}$         & $84.13_{\pm 0.56}$        & $\bs{85.09_{\pm 0.59}}$  \\
        ogbn-arxiv      & $52.80_{\pm 0.19}$       & $53.81_{\pm 0.23}$         & $70.48_{\pm 0.18}$        & $68.77_{\pm 0.06}$         & $70.50_{\pm 0.11}$       & $69.89_{\pm 0.59}$        & $\ul{71.04_{\pm 0.27}}$   & $66.60_{\pm 0.53}$        & $\bs{71.48_{\pm 0.21}}$   &    $68.50_{\pm 0.17}$         & $59.28_{\pm 2.22}$        & $70.73_{\pm 0.17}$ \\
		\midrule
		{\bf Avg. Rank}	& 9.21					   & 7.86					    & 6.79					    & 7.69					     & 3.50					    & 7.86                      & 7.57					    & 10.50                     & 3.50				        & 4.36                       & 7.29                      & 1.57\\
		\bottomrule
	\end{tabular}
	}
\end{table*}

\paragraph{Settings and Results.}
For all datasets except ogbn-arxiv, we generate 10 random splits with a consistent  train/valid/test ratio of 60\%/20\%/20\%.
For the ogbn-arxiv dataset, we generate 10 random splits using the \texttt{get\_idx\_split()} function provided by the official package.
We set the hidden units to 64 for all datasets.
We then tune the hyper-parameters 100 times for each model on each dataset using the TPESampler from the optuna package.
The ranges of the hyper-parameters are listed below:
\begin{itemize}
    \item \texttt{learning\_rate} for all models: \{1e-3, 5e-3, 1e-2, 5e-2\};
    \item \texttt{weight\_decay} for all models: \{0.0, 1e-5, 5e-5, 1e-4, 5e-4, 1e-3, 5e-3, 1e-2\};
    \item \texttt{dropout} for all models: \{0.0, 0.1, 0.3, 0.5\};
    \item \texttt{num\_layers} for all models: \{2, 4, 8\};
    \item \texttt{alpha} for APPNP, GCNII and MGNN: \{0.00, 0.01, $\cdots$, 0.99, 1.00\};
    \item \texttt{beta} for MGNN: \{0.00, 0.01, $\cdots$, 0.99, 1.00\};
    \item \texttt{theta} for GCNII and MGNN: \{0.5, 1.0, 1.5\};
    \item \texttt{mu} for $^p$GNN: \{0.00, 0.01, $\cdots$, 0.99, 1.00\};
    \item \texttt{p} for $^p$GNN: \{0.5, 1.0, 1.5, 2.0\};
    \item \texttt{initial}: for MGNN: \{`Linear', `MLP'\};
    \item \texttt{attention} for MGNN: \{`concat', 'bilinear'\}.
\end{itemize}

We train each model for a maximum of 1500 epochs, with early stopping set to 100, on each dataset and report the results after hyper-tuning.
The results of the supervised node classification experiments are presented in Table \ref{tab:supervised}.
We report the mean accuracy and the standard deviation based on 10 random splits generated as previously described.
Our MGNN model generally performs well across all datasets, and outperforms all baselines in 10 out of 14 datasets.
We also include the average rank of each model across all datasets.
Our MGNN model achieves an average rank of 1.57, which is the highest among all models.

\paragraph{Ablation.}
Comparing the convolution rule of GCNII,
\begin{equation}
    \resizebox{\linewidth}{!}{
    $\bs{Z}^{(k + 1)} = \sigma(((1 - \alpha)\tilde{\bs{A}}\bs{Z}^{(k)} + \alpha \bs{Z}^{(0)})(\gamma^{(k)}\bs{W}^{(k)} + (1 - \gamma^{(k)})\bs{I})),$
    }
\end{equation}
with that of our MGNN model (Equation \ref{eqn:MGNN-update}), we can conclude that our MGNN model is a generalization of the GCNII model.
In other words, the GCNII model can be considered as an ablation study of the MGNN model.

\subsection{Graph Regression}
To demonstrate the overall effectiveness of our MGNN model in graph regression tasks, we conducted experiments using the GCN model, GAT model, GIN model, PointNet model, and our MGNN model on the ZINC-subset dataset.
Due to space constraints, we provide the detailed experiment settings in Appendix \ref{app:regression}.
A summary of the results can be found in Table \ref{tab:regression}.
\begin{table}[h]
    \centering
    \caption{The results of the graph regression experiments.}
    \label{tab:regression}
    \resizebox{\linewidth}{!}{
    \begin{tabular}{ccccc}
        \toprule
        GCN & GAT & GIN & PointNet & MGNN \\
        \midrule
        $0.6143_{\pm 0.0200}$ & $0.6458_{\pm 0.0647}$ & $\bs{0.4410_{\pm 0.0065}}$ & $0.5293_{\pm 0.0138}$ & $\ul{0.4751_{\pm 0.0112}}$\\
        \bottomrule
    \end{tabular}
    }
\end{table}

\section{Conclusion}
In this paper, we draw inspiration from the fact that typical classifiers, such as linear layers and MLPs, are not sensitive to congruent transformations. 
Based on this, we discuss the sufficient conditions for a spatial-universal GNN. 
To strike a balance between efficiency and expressive power, we propose a more relaxed model called MGNN. 
The optimization objective of MGNN is primarily focused on minimizing the raw stress arising from the distance geometry problem and the multidimensional scaling problem.
We conducted extensive experiments on both synthetic datasets and real-world datasets. 
In the experiments involving the arrangement of nodes based on a given metric matrix, we observed that our MGNN model strives to separate blocks of nodes, particularly in the case of homophilic graphs. 
Furthermore, in node classification experiments, our MGNN model demonstrated superior performance in real-world applications.
We hope that our model will provide new insights into spatial GNNs and inspire further explorations in this area.

\begin{acks}
This research was supported in part by National Natural Science Foundation of China (No. U2241212, No. 61972401, No. 61932001, No. 61832017), by the major key project of PCL (PCL2021A12), by Beijing Natural Science Foundation (No. 4222028), by Beijing Outstanding Young Scientist Program No.BJJWZYJH012019100020098, by Alibaba Group through Alibaba Innovative Research Program, and by Huawei-Renmin University joint program on Information Retrieval. We also wish to acknowledge the support provided by Engineering Research Center of Next-Generation Intelligent Search and Recommendation, Ministry of Education. Additionally, we acknowledge the support from Intelligent Social Governance Interdisciplinary Platform, Major Innovation \& Planning Interdisciplinary Platform for the “Double-First Class” Initiative, Public Policy and Decision-making Research Lab, Public Computing Cloud, Renmin University of China. 
\end{acks}
\newpage

\bibliographystyle{ACM-Reference-Format}
\balance
\bibliography{MGNN}

\newpage
\appendix
\section{Proof of Theorem \ref{thm:groups-relation}}
\label{app:prf-group-relation}
\begin{proof}
	We will prove this theorem in three parts.
	\begin{itemize}
		\item $\mathrm{Iso}(n, d; V^2)\le \mathrm{Iso}(n, d; E)$. 
			It is evident that every congruent transformation $\mathcal{T}\in \mathrm{Iso}(n, d; V^2)$ is also equivalent, meaning $\mathcal{T}\in \mathrm{Iso}(n, d; E)$, since $E\subseteq V^2$.
		\item $\mathrm{E}(d)\le \mathrm{Iso}(n, d; V^2)$.
			For an orthogonal matrix $\bs{Q}$, we can verify that $$(\bs{ZQ})(\bs{ZQ})^\top = \bs{ZQQ}^\top\bs{Z}^\top = \bs{ZZ}^\top,$$
            which implies $M(\bs{Z}) = M(\bs{ZQ})$.
			For a translation vector $\bs{s}\in \mathbb{R}^d$, we can verify that for all $i, j \in \{1, 2, \cdots, n\}$, 
            $$M(\bs{Z}+\bs{1}_n\bs{s}^\top)_{ij} =\Vert(\bs{Z}_{i:} + \bs{s}^\top) - (\bs{Z}_{j:} + \bs{s}^\top)\Vert_2 = M(\bs{Z})_{ij},$$ hence $M(\bs{Z} + \bs{1}_n\bs{s}^\top) = M(\bs{Z})$.
			Therefore, for any $\mathcal{T}\in \mathrm{E}(d)$, we have $$M(\mathcal{T}(\bs{Z})) = M(\bs{ZQ}+\bs{1}_n\bs{s}^\top) = M(\bs{Z}),$$ which means $\mathcal{T}\in \mathrm{Iso}(n, d; V^2)$.
		\item $\mathrm{Iso}(n, d; V^2)\le \mathrm{E}(d)$. 
			Suppose we have two embeddings $\bs{X}\in \mathbb{R}^{n\times d}$ and $\bs{Y}=\mathcal{T}(\bs{X})$ for some $\mathcal{T}\in \mathrm{Iso}(n, d; V^2)$.
            We need to demonstrate that $\mathcal{T}'$ can be expressed as is the composition of an orthogonal transformation and a translation transformation.
			
            First, we translate $\bs{X}_{1:}$ and $\bs{Y}_{1:}$ to the origin by defining $\bs{X}'=\bs{X}-\bs{1}_n\bs{X}_{1:}$ and $\bs{Y}'=\bs{Y}-\bs{1}_n\bs{Y}_{1:}$.
			The metric matrix is preserved under translation, i.e., $M(\bs{Y}') = M(\bs{Y}) = M(\bs{X}) = M(\bs{X'})$.
            Now, for $i\in \{1, 2, \cdots, n\}$, we have $$\Vert\bs{X}'_{i:}\Vert = \Vert \bs{X}'_{i:} - \bs{0}\Vert = \Vert \bs{X}'_{i:} - \bs{X}_{1:}'\Vert = M(\bs{X}')_{i1} = M(\bs{Y}')_{i1}.$$
            For $\bs{Y}'$, we have:
            $$\Vert\bs{Y}'_{i:}\Vert = M(\bs{Y}')_{i1}.$$
            
			Let us denote $\bs{Y}'$ as $\bs{Y}' = \mathcal{T}'(\bs{X}')$, and we will now show that $\mathcal{T}'$ is an orthogonal transformation.
   
            We first show that $\mathcal{T}'$ preserves the inner product, i.e., \\$\bs{Y}'(\bs{Y}')^\top = \bs{X}'(\bs{X}')^\top$.
            Note that for $i, j \in \{1, 2, \cdots, n\}$, we have 
            \begin{equation}
                \begin{split}
                    {\bs{X}'_{i:}}^\top \bs{X}'_{j:} 
                    & = \dfrac{1}{2}(\Vert \bs{X}'_{i:}\Vert^2 + \Vert \bs{X}'_{j:}\Vert^2 - \Vert \bs{X}'_{i:} - \bs{X}'_{j:}\Vert^2) \\
                    & = \dfrac{1}{2}(M(\bs{X}')_{i1} + M(\bs{X}')_{j1} - M(\bs{X}')_{ij}),
                \end{split}
            \end{equation}
            and similarly, 
            \begin{equation}
                \begin{split}
                    {\bs{Y}'_{i:}}^\top \bs{Y}'_{j:} 
                    & = \dfrac{1}{2}(\Vert \bs{Y}'_{i:}\Vert^2 + \Vert \bs{Y}'_{j:}\Vert^2 - \Vert \bs{Y}'_{i:} - \bs{Y}'_{j:}\Vert^2) \\
                    & = \dfrac{1}{2}(M(\bs{Y}')_{i1} + M(\bs{Y}')_{j1} - M(\bs{Y}')_{ij}).
                \end{split}
            \end{equation}
            Since $M(\bs{X}') = M(\bs{Y}')$, we have
			$\left((\bs{Y}')(\bs{Y}')^\top\right)_{ij} = \bs{Y}'_{i:}(\bs{Y}')_{j:}^\top = \bs{X}'_{i:}(\bs{X}'_{j:})^\top = \left((\bs{X}')(\bs{X}')^\top\right)_{ij}$,
			thus $\bs{Y}'(\bs{Y}')^\top = \bs{X}'(\bs{X}')^\top$.
   
			We then demonstrate that $\mathcal{T}'$ is linear. 
			For any vector $\bs{x}$, $\bs{y}\in \mathbb{R}^d$ and scalars $a$, $b$ in $\mathbb{R}$, let $\bs{z} = \mathcal{T}'(a\bs{x} + b\bs{y}) - a\mathcal{T}'(\bs{x}) - b\mathcal{T}'(\bs{y})$.
			We can then verify that $\bs{z}^\top\bs{z} = 0$ through some calculations:
            \begin{align*}
                \bs{z}^\top\bs{z} 
                & = \mathcal{T}'(a\bs{x} + b\bs{y})^\top\mathcal{T}'(a\bs{x} + b\bs{y}) + a^2\mathcal{T}'(\bs{x})^\top\mathcal{T}'(\bs{x}) \\
                & \qquad + b^2\mathcal{T}'(\bs{y})^\top\mathcal{T}'(\bs{y}) - 2a\mathcal{T}'(a\bs{x} + b\bs{y})^\top\mathcal{T}'(\bs{x}) \\
                & \qquad - 2b\mathcal{T}'(a\bs{x}+b\bs{y})^\top\mathcal{T}'(\bs{y}) + 2ab\mathcal{T}'(\bs{x})^\top\mathcal{T}'(\bs{y}) \\
                & = (a\bs{x}+b\bs{y})^\top(a\bs{x}+b\bs{y}) + a^2\bs{x}^\top\bs{x} + b^2\bs{y}^\top\bs{y} \\ 
                & \qquad - 2a(a\bs{x}+b\bs{y})^\top\bs{x} - 2b(a\bs{x}+b\bs{y})^\top\bs{y} + 2ab\bs{x}^\top\bs{y} \\
                & = 2a^2\bs{x}^\top\bs{x} + 2b^2\bs{y}^\top\bs{y} + 2ab\bs{x}^\top\bs{y} \\
                & \qquad - 2a^2\bs{x}^\top\bs{x} - 2ab\bs{x}^\top\bs{y} - 2ab\bs{x}^\top\bs{y} - 2b^2\bs{y}^\top\bs{y} + 2ab\bs{x}^\top\bs{y} \\
                & = 0.
            \end{align*}
			Consequently, we find that $\bs{z} = \bs{0}$, which implies $\mathcal{T}'(a\bs{x} + b\bs{y}) = a \mathcal{T}'(\bs{x}) + b\mathcal{T}'(\bs{y})$. Therefore, the mapping $\mathcal{T}'$ is linear.
   
            Since $\mathcal{T}'$ is linear, it can be represented by a matrix $\bs{Q}$ (with respect to the standard orthonormal basis), such that $\mathcal{T}'(\bs{x}) = \bs{Qx}$.
            Hence, $\Vert \mathcal{T}'(\bs{x})\Vert = \Vert \bs{x}\Vert$ if and only if $(\bs{Qx})^\top\bs{Qx} = \bs{x}^\top\bs{x}$, which further leads to $\bs{Q}^\top\bs{Q} = \bs{I}$.
			From this, we can conclude that $\mathcal{T}'$ is an orthogonal transformation.

            Therefore, $\mathcal{T}(\bs{X}) = \bs{Q}(\bs{X} - \bs{1}_n\bs{X}_{1:}) + \bs{1}_n\bs{Y}_{1:} = \bs{Q}\bs{X} - \bs{Q}\bs{1}_n\bs{X}_{1:} + \bs{1}_n\bs{Y}_{1:}$, which implies $\mathcal{T} \in \mathrm{E}(d)$.
	\end{itemize}
	This concludes the proof.
\end{proof}

\section{Proof of Theorem \ref{thm:mlp-congruent-insensitive}}
\label{app:prf-mlp-congruent-insensitive}
\begin{proof}
	Since the composition of orthogonal and translation transformations are the only types of congruent transformations in $\mathbb{R}^d$ (according to Theorem \ref{thm:groups-relation}), 
		we can express $\bs{Z}_2$ as $\bs{Z}_2 = \bs{Z}_1\bs{Q}+\bs{1}_{n}\bs{s}^\top$, where $\bs{Q}\in \mathrm{O}(d)$ is an orthogonal matrix, $\bs{1}_{n}\in \mathbb{R}^n$ is an all-one vector, and $\bs{s}\in \mathbb{R}^d$ is a translation vector.
	Thus, for any $\mathrm{MLP}_M$ with parameters $\bs{W}_M^{(0)}, \cdots, \bs{W}_M^{(L-1)}$ and $\bs{b}_M^{(0)}, \cdots, \bs{b}_M^{(L-1)}$, we can construct another $\mathrm{MLP}_{N}$ with parameters $\bs{W}_N^{(0)}, \bs{W}_M^{(1)}\cdots, \bs{W}_M^{(L-1)}$ and $\bs{b}_N^{(0)}, \bs{b}_M^{(1)}\cdots, \bs{b}_M^{(L-1)}$, where $\bs{W}_{N}^{(0)} = \bs{Q}^{-1}\bs{W}_{M}^{(0)}$ and $\bs{b}_{N}^{(0)} = \bs{b}_{M}^{(0)}-(\bs{Q}^{-1}\bs{W}_{M}^{(0)})^\top\bs{s}$.
	It is easy to verify that \\$\mathrm{MLP}_{M}(\bs{Z}_1) = \mathrm{MLP}_{N}(\bs{Z}_2)$, and the theorem follows.
\end{proof}

\section{Proof of Theorem \ref{thm:invariant}}
\label{app:prf-invariant}
\begin{proof}
	Expanding the left side of the equation and using the definition of $\mathrm{Iso}(E)$, we have:
	\begin{align*}
		E_p(\sigma(\bs{Z}); \bs{M}, E) 
		& = \sum_{(i, j)\in E}\frac{1}{2}\left(\left\Vert\sigma(\bs{Z})_{i:}-\sigma(\bs{Z})_{j:}\right\Vert_2 - \bs{M}_{ij}\right)^2 \\
		& = \sum_{(i, j)\in E}\frac{1}{2}\left(\left\Vert\bs{Z}_{i:}-\bs{Z}_{j:}\right\Vert_2-\bs{M}_{ij}\right)^2 \\
		& = E_p(\bs{Z}; \bs{M}, E).
	\end{align*}
    Therefore, the theorem follows.
\end{proof}

\section{Visualization Results of SGC and APPNP}
\label{app:vis-results}
\begin{figure*}[ht]
    \centering
    \includegraphics[width = \textwidth]{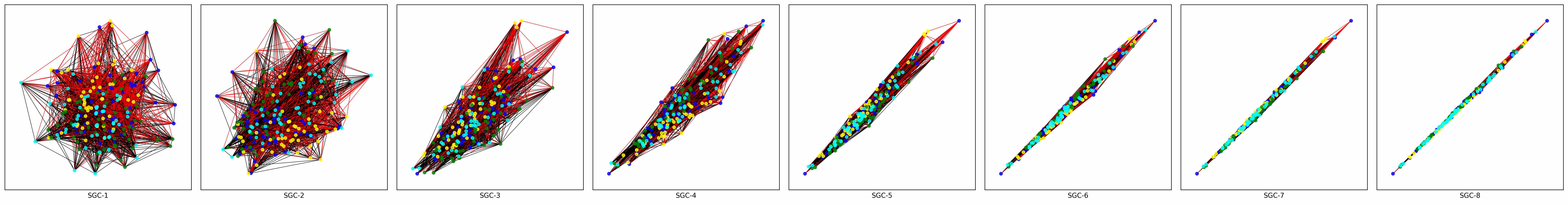}
    \caption{The results of the homophilic SBM graph of SGC layers.}
    \label{fig:SGC-homo}
\end{figure*}

\begin{figure*}[ht]
    \centering
    \includegraphics[width = \textwidth]{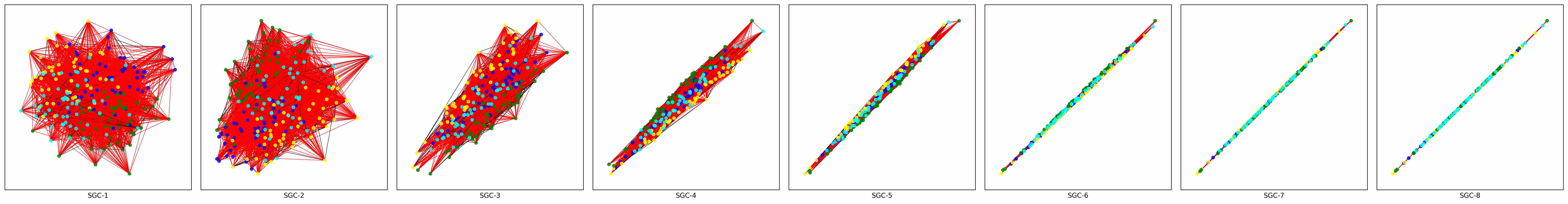}
    \caption{The results of the heterophilic SBM graph of SGC layers.}
    \label{fig:SGC-hetero}
\end{figure*}

\begin{figure*}[ht]
    \centering
    \includegraphics[width = \textwidth]{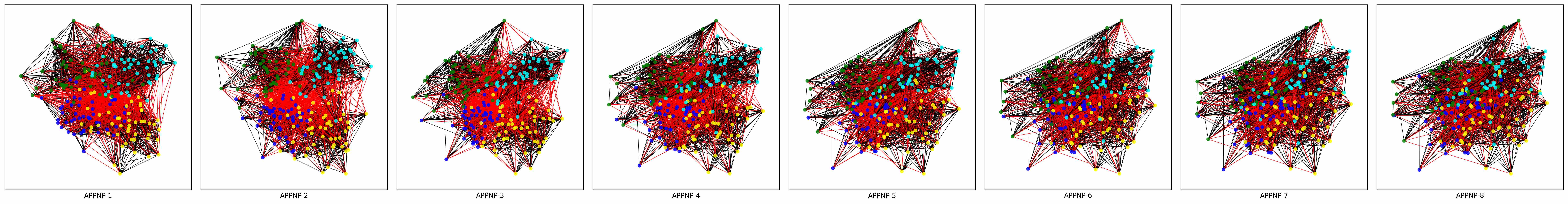}
    \caption{The results of the homophilic SBM graph of APPNP layers.}
    \label{fig:APPNP-homo}
\end{figure*}

\begin{figure*}[ht]
    \centering
    \includegraphics[width = \textwidth]{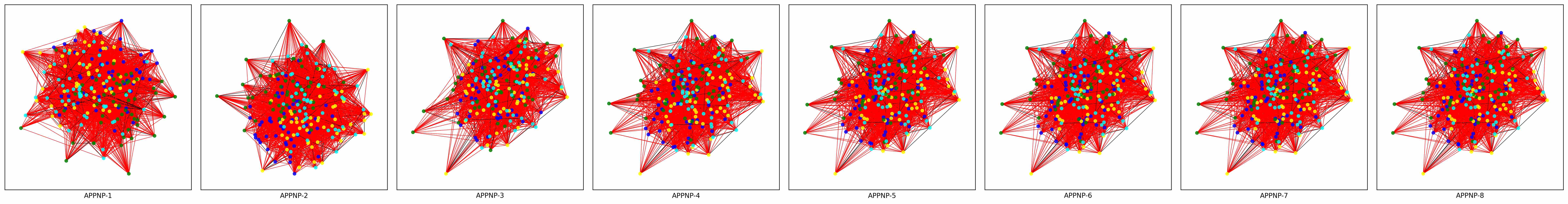}
    \caption{The results of the heterophilic SBM graph of APPNP layers.}
    \label{fig:APPNP-hetero}
\end{figure*}

We present the visualization results of the SGC layers (Figure \ref{fig:SGC-homo} and Figure \ref{fig:SGC-hetero}) and APPNP layers (Figure \ref{fig:APPNP-homo} and Figure \ref{fig:APPNP-hetero}) for the two synthetic graphs described earlier.

\section{Graph Regression}
\label{app:regression}
\paragraph{Settings.}
We utilized the publicly available train/validation/test split provided by the PyTorch Geometric Library.
For all datasets, we set the number of hidden units to 64.
Subsequently, we conduct hyperparameter tuning for each model on each dataset by employing the TPESampler from the optuna package. 
This process is repeated 100 times.
Below, we present the ranges of the hyperparameters:
\begin{itemize}
    \item \texttt{learning\_rate} for all models: \{1e-3, 5e-3, 1e-2, 5e-2\};
    \item \texttt{weight\_decay} for all models: \{0.0, 1e-5, 5e-5, 1e-4, 5e-4, 1e-3, 5e-3, 1e-2\};
    \item \texttt{dropout} for all models: \{0.0, 0.1, 0.3, 0.5\};
    \item \texttt{num\_layers} for all models: \{2, 4, 8\};
    \item \texttt{alpha} for MGNN: \{0.00, 0.01, $\cdots$, 0.99, 1.00\};
    \item \texttt{beta} for MGNN: \{0.00, 0.01, $\cdots$, 0.99, 1.00\};
    \item \texttt{theta} for MGNN: \{0.5, 1.0, 1.5\};
    \item \texttt{initial}: for MGNN: \{`Linear', `MLP'\};
    \item \texttt{attention} for MGNN: \{`concat', 'bilinear'\}.
\end{itemize}
Due to limited experiment time, we limit the training of each model to a maximum of 500 epochs. 
A batch size of 512 is utilized, and early stopping is implemented after 50 epochs on each dataset. 
Following hyperparameter tuning, we report the results.
To enhance reliability, we repeat this process five times and calculat the mean MAE (Mean Absolute Error) and standard deviation for each model.

\paragraph{Analysis of the Results.}
We observed in our experiments that our MGNN model generally required more epochs to converge better, so the small epochs and insufficient training may have hindered its performance. 
In addition, although the GIN model outperforms other models under our graph regression experiment setting, it generally performs poorly in the node classification task. 
Therefore, our MGNN model has better overall performance than the GIN model. 
It is also important to emphasize that our MGNN model only relies on the {\bfseries partial} metric information defined on the edges, so it cannot be directly compared to those transformers or global GNNs that utilize the full (global or all-pair) metric information.

\end{document}